\documentclass[10pt,twocolumn,letterpaper]{article}
\usepackage{iccv}

\usepackage[T1]{fontenc}
\usepackage{gensymb}
\usepackage{xspace}
\usepackage{listings}
\usepackage{tikz}
\usepackage[most]{tcolorbox}

\definecolor{stringcolor}{rgb}{0.75,0.2,0.2} % Strings (red)
\definecolor{keycolor}{rgb}{0.1,0.1,0.7}     % Keys (blue)
\definecolor{numbercolor}{rgb}{0.2,0.5,0.2}  % Numbers (green)

\lstdefinelanguage{json}{
    basicstyle=\ttfamily\footnotesize,      
    breaklines=true,              
    upquote=true,                 
    showstringspaces=false,       
    string=[db]{"},
    stringstyle=\color{black},
    morestring=[s][\color{black}]{\ \ "}{":},
    keywordstyle=\color{blue},
    keywords={true,false,null},
    literate=
     *{0}{{{\color{red}0}}}{1}
      {1}{{{\color{red}1}}}{1}
      {2}{{{\color{red}2}}}{1}
      {3}{{{\color{red}3}}}{1}
      {4}{{{\color{red}4}}}{1}
      {5}{{{\color{red}5}}}{1}
      {6}{{{\color{red}6}}}{1}
      {7}{{{\color{red}7}}}{1}
      {8}{{{\color{red}8}}}{1}
      {9}{{{\color{red}9}}}{1}
      {.}{{{\color{red}.}}}{1}
      {:}{{{\bfseries\color{green!50!black}{:}}}}{1}
      {,}{{{\bfseries\color{green!50!black}{,}}}}{1}
      {\{}{{{\bfseries\color{green!50!black}{\{}}}}{1}
      {\}}{{{\bfseries\color{green!50!black}{\}}}}}{1}
      {[}{{{\bfseries\color{green!50!black}{[}}}}{1}
      {]}{{{\bfseries\color{green!50!black}{]}}}}{1},            
}

\newtcblisting{roundedlisting}{
  colback=lightgray!10, 
  colframe=lightgray,  
  listing only,       
  arc=2mm,            
  boxrule=0.25mm,      
  left=1pt, right=1pt, top=0pt, bottom=0pt, 
  enhanced,        
  listing options={language=json}   
}

\makeatletter
\renewcommand{\paragraph}{%
  \@startsection{paragraph}{4}%
  {\z@}{-0.5em}{-0.5em}%
  {\normalfont\normalsize\bfseries}%
}
\makeatother

\newcommand{\method}{SynCity\xspace}

\definecolor{iccvblue}{rgb}{0.21,0.49,0.74}
\usepackage[pagebackref,breaklinks,colorlinks,allcolors=iccvblue]{hyperref}

\def\paperID{{*****}}
\def\confName{ICCV}
\def\confYear{2025}

\title{SynCity: Training-Free Generation of 3D Worlds}

\author{
Paul Engstler$^*$ 
\and
Aleksandar Shtedritski$^*$ 
\and
Iro Laina
\and
Christian Rupprecht
\and
Andrea Vedaldi \\
Visual Geometry Group, University of Oxford\\
{\tt\small \{paule,suny,iro,chrisr,vedaldi\}@robots.ox.ac.uk}
}

\begin{document}

\twocolumn[
\maketitle
\begin{center}
\fcolorbox{white}{white}{
\includegraphics[width=0.97\textwidth]{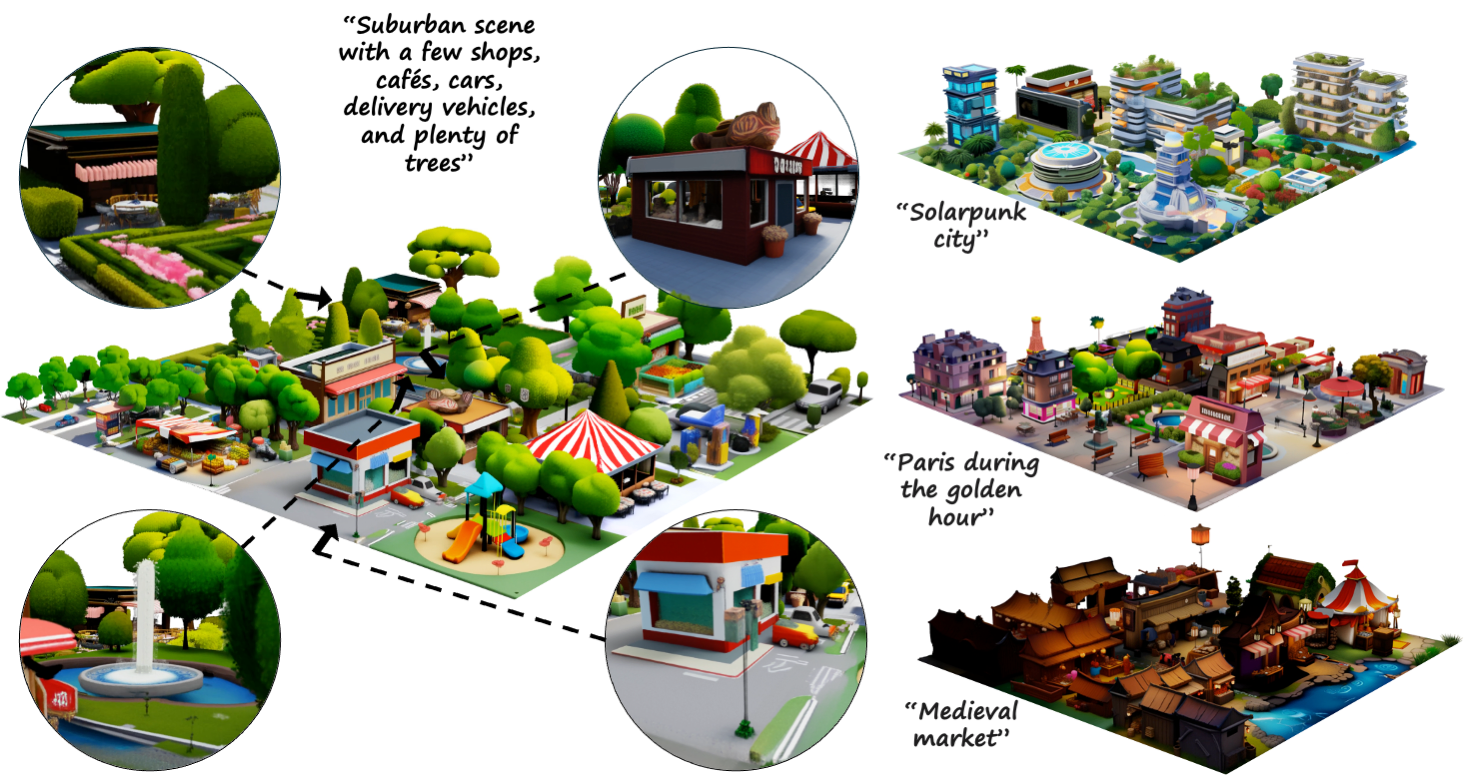}
}
\end{center}
\vspace{-1em}
\captionof{figure}{
We introduce \textbf{SynCity}, a novel method that can generate from a prompt complex and immersive 3D worlds that can be navigated freely.
Our method is training-free and leverages powerful language, 2D and 3D generators via novel prompt engineering strategies.}%
\label{fig:splash}
\vspace{2em}
]

\def\thefootnote{*}\footnotetext{Equal contribution} 
\begin{center}
\bfseries\large Abstract
\vspace{-0.5em}
\end{center}

\textit{
We address the challenge of generating 3D worlds from textual descriptions. 
We propose \method, a training- and optimization-free approach, which leverages the geometric precision of pre-trained 3D generative models and the artistic versatility of 2D image generators to create large, high-quality 3D spaces. 
While most 3D generative models are object-centric and cannot generate large-scale worlds, we show how 3D and 2D generators can be combined to generate ever-expanding scenes.
Through a tile-based approach, we allow fine-grained control over the layout and the appearance of scenes.
The world is generated tile-by-tile, and each new tile is generated within its world-context and then fused with the scene.
\method generates compelling and immersive scenes that are rich in detail and diversity.
}    
\section{Introduction}%
\label{sec:intro}

We consider the problem of generating 3D worlds from textual descriptions.
Generating 3D content, for example, for video games, virtual reality, special effects and simulation, is highly laborious and time-consuming.
When it comes to generating entire 3D scenes, much of the content is not of particular artistic value, and its manual creation, which is still necessary, may be seen as a waste of human resources, talent and creativity.
Generative models can help to reduce or even remove this burden by largely automating many of these mundane tasks.

The advent of modern generative AI has significantly impacted 3D content generation and promises to reduce the cost of its production dramatically.
DreamFusion~\cite{poole23dreamfusion:} was among the first to co-opt state-of-the-art diffusion-based 2D image generators~\cite{rombach22high-resolution} to create 3D objects.
The area has since matured significantly by fine-tuning 2D image generators to produce multiple consistent views of an object~\cite{shi24mvdream:,szymanowicz23viewset,melas-kyriazi24im-3d,gao24cat3d:} and by learning few-view 3D reconstruction networks~\cite{li24instant3d:,xu24grm:}.
More recently, the focus has shifted to methods that learn a 3D latent space~\cite{zhang233dshape2vecset:,deemos24rodin,li25triposg:,li24craftsman:,xiang24structured}, which can then be sampled to generate 3D objects.
Because the latent space directly encodes the 3D structure of the object rather than its 2D appearance, these methods can generate much more accurate and regular geometry.

Despite their advantages, 3D generative methods have been so far mostly limited to generating individual objects.
However, the most promising usage of 3D generative AI is the construction of entire virtual worlds, as this is where automation can make by far the most difference.
There is ample literature on generating 3D scenes from textual or image prompts.
Most such methods are image-based, 
and progressively reconstruct larger scene regions by expanding from an initial image~\cite{rockwell21pixelsynth:, hollein23text2room:, chung23luciddreamer:, yu23wonderjourney:, zhang23text2nerf:, liu21infinite, fridman23scenescape:, ouyang23text2immersion:, engstler25invisible}, combining depth prediction, image and depth outpainting, and 3D reconstruction using NeRF~\cite{mildenhall20nerf:} or 3D Gaussian Splatting~\cite{kerbl233d-gaussian}.
The main advantage of these approaches is that they can leverage powerful 2D image generator models to create the first and subsequent views of the scene.
These 2D generators allow the overall system to understand complex textual prompts and generate corresponding 3D scenes with good artistic quality.
However, it is difficult for these approaches to maintain a coherent 3D structure across a large scene.
For example, while the reconstructed scene may envelop the observer in a 360\textdegree{} manner, it is generally not possible to `walk' into the scene for more than a few steps.
This is the case even for state-of-the-art implementations like the one of World Labs~\cite{worldlabs24generating}, a company specialized in SpatialAI\@.

\begin{figure*}[t]
\centering
\includegraphics[width=0.98\textwidth, trim={0 1.5cm 0 0},clip]{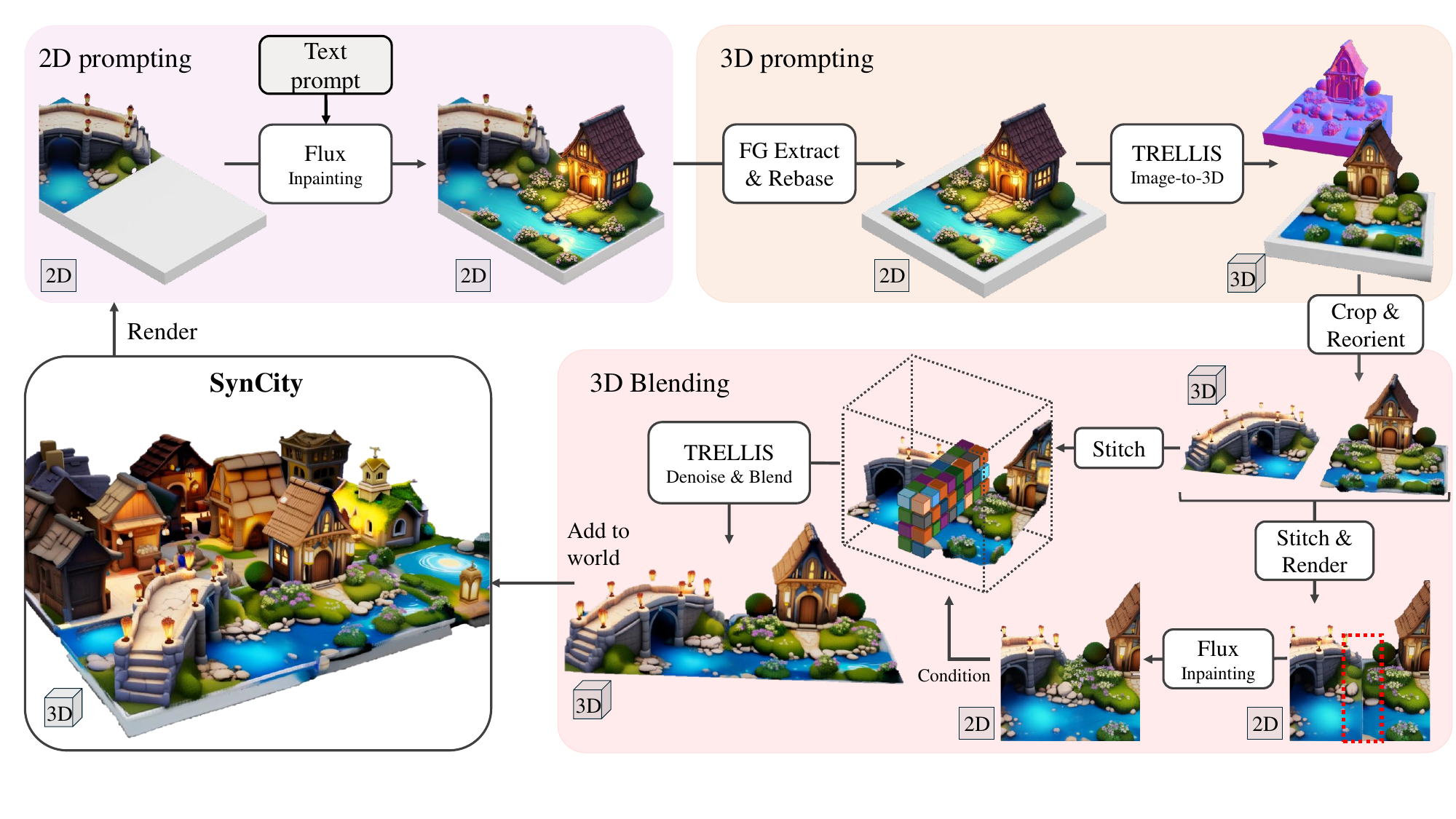}\caption{\textbf{Overview of SynCity.}
2D prompting: To generate a new tile, we first render a view of where that tile should be placed, including context from neighbouring tiles.
3D prompting: We extract the new tile image and construct an image prompt for TRELLIS by adding a wider base under the tile.
3D blending: The 3D model that TRELLIS outputs is usually not well blended with the rest of the scene.
To address that,  we render a view of the new tile next to each neighbouring tile, and inpaint the region between the two with an image inpainting model.
Next, we condition using that well-blended view to refine the region between the two 3D tiles.
Finally, the new, blended, tile is added to the world.}%
\label{fig:main}
\end{figure*}

A challenge with extending scenes beyond these `3D bubbles' is that it is difficult for image-based methods to maintain consistency incrementally without drifting.
We argue that 3D generative models might be preferable as they can regularize and constrain the reconstructed geometry, including hallucinating shape and textures in regions \emph{behind} the visible sides of objects.
This is clearly shown in prior works like BlockDiffusion~\cite{wu24blockfusion:} and LT3SD~\cite{meng24lt3sd:}, where large coherent spaces can be generated.
However, these methods are limited in the quality and diversity of the generated scenes as it is difficult to train 3D generative models directly for scene generation.
In particular, unlike their view-based peers, these methods do not build on an image generator. They thus cannot benefit from the artistic quality and ability to interpret complex textual prompts that come from pre-training 2D generators on billions of images.

In this work, we thus seek to build on 3D generative models while still building on the strengths of 2D image generators to generate large, high-quality 3D spaces that can be navigated freely (\cref{fig:splash}).
First, we note that while 3D models like TRELLIS~\cite{xiang24structured} are trained for object-level reconstruction, they can reconstruct fairly complex local compositions of multiple objects.
Borrowing ideas from video game world building, we show in particular that TRELLIS can effectively generate, if not an entire world, at least a \emph{tile} representing a local region of the world.
We show, in particular, how to prompt the model with an `isometric' view of the tile and then generate the tile in 3D.

Given this basic capability, we then look at the problem of generating a large scene by generating and stitching together multiple tiles.
We build on a text-to-image generator (Flux~\cite{flux}) and propose a novel way of prompting that \emph{stabilizes} it to consistently produce tiles with a similar isometric framing.
In this manner, we encourage the tiles' framing to be stable and compatible between different tiles, making it easier to stitch them together.

To ensure tiles to fit together appropriately, we propose two mechanisms. 
First, we encourage consistency in appearance by using previously generated tiles to draw context for the image generator, where each new tile \emph{inpaints} a missing region in a 2D isometric view of the scene.
Secondly, we enforce geometric consistency by blending the 3D representations of neighbouring tiles using the 3D generative model.

\section{Related Work}%
\label{s:related}

\paragraph{Novel view synthesis for scenes.}
Expanding an image beyond its boundaries has been a long-standing task in computer vision. Early methods that sought to expand object-centric scenes rely on layer-structured representations~\cite{tulsiani18layer-structured, mildenhall19local, srinivasan2019pushing, tucker2020single, li21mine:, shih203d-photography}, which disregard the scene's true geometry. SynSin~\cite{wiles20synsin:} is a pivotal work, where image features are projected and used as conditioning to generate novel views, achieving geometric and semantic consistency.
ZeroNVS~\cite{sargent23zeronvs:} introduces high-quality results with fine-grained control of the camera but remains object-centric. GenWarp~\cite{seo24genwarp:} integrates semantic information through cross-attention when generating a novel view.

The major challenges for these methods remain semantic drift and object permanence. To obtain an explicit 3D representation, the generated views need to be transferred into such a representation, e.g., NeRF~\cite{mildenhall20nerf:} or Gaussians~\cite{kerbl233d-gaussian, huang20242d}, where any geometric conflicts would need to be resolved.

\paragraph{Image projection-based scene generation.}
A different line of work follows the paradigm of building the 3D representation of a scene sequentially using 2D image generation models~\cite{rockwell21pixelsynth:, wang2023prolificdreamer, hollein23text2room:, fridman23scenescape:, lei23rgbd2:, xiang233d-aware, chung23luciddreamer:, yu23wonderjourney:, zhang23text2nerf:, ouyang23text2immersion:}. Most of them employ an image generation model to outpaint the existing scene using pre-defined camera poses. The results are then fused in 3D with depth prediction models. Text2Room~\cite{hollein23text2room:} generates meshes of indoor scenes. As the scene is clearly delimited by the bounds of the mesh, it can be freely explored.
LucidDreamer~\cite{chung23luciddreamer:} and Text2Immersion~\cite{ouyang23text2immersion:} go beyond indoor scenes, but their generated scenes reveal geometric inconsistencies when stepping away from the camera poses used to generate the scene. Invisible Stitch~\cite{engstler25invisible} addresses this issue by inpainting depth (rather than naively aligning it) and RealmDreamer~\cite{shriram25realmdreamer:} proposes multiple optimization losses to refine the generated scene. Despite these improvements, the resulting scenes still suffer from geometric artifacts and remain rather small.
WonderJourney~\cite{yu23wonderjourney:} introduces novel ideas for depth fusion, such as grouping objects at similar disparity to planes and sky depth refinement, enabling large `scene journeys', where independent representations are built between scene `keyframes', but these are not merged into one coherent scene. WonderWorld~\cite{yu2024wonderworld} leverages these improvements to build a single scene, allowing interactive updates, but the true extent of the generated scenes remains limited. Other works use panoramas~\cite{wang24perf:, stan23ldm3d:} or implicit representations~\cite{bautista22gaudi:, sargent23zeronvs:} but the freedom of movement remains constricted.

\paragraph{Procedural scene generation.}
Further methods permit long-range fly-overs over nature~\cite{liu21infinite, li2022infinitenature, cai23diffdreamer:, chen2023scenedreamer, chai2023persistent} or cities~\cite{lin2023infinicity, xie2024citydreamer, shang2024urbanworld}. These usually generate procedural unbounded images (e.g., the terrain make-up or a city layout). While those methods create realistic-looking images, they are often monotonous as the methods are domain-specific and thus highly constrained in the variety they can generate.

\paragraph{3D scene generation.}
Instead of merely generating images of a scene or outpainting it only in 2D, further methods generate the representation directly. Set-the-Scene~\cite{cohen-bar23set-the-scene:} adds a layer of control to the layout of NeRF scenes by defining object proxies. BlockFusion~\cite{wu24blockfusion:} learns a network to auto-regressively diffuse small blocks to extend a mesh. A 2D layout conditioning is used to control the generation process, allowing users to generate scenes of rooms, a village, and a city. While the method allows building large-scale scenes, the variety of the objects it generates is severely limited as it requires domain-specific 3D training data. Furthermore, it generates untextured meshes. LT3SD~\cite{meng24lt3sd:} learns a diffusion model that generates 3D environments in a patch-by-patch and coarse-to-fine fashion. However, this method is only trained to produce indoor scenes.

At the same time, the synthesis of complex, high-fidelity objects has been enabled by the rapid progress in the fields of text-to-3D and image-to-3D generation~\cite{lin22magic3d:, liu23zero-1-to-3:,qian23magic123:, shi24mvdream:, raj23dreambooth3d:, tang24lgm:, li25triposg:,zhang2024clay, li24craftsman:,xiang24structured, zhao2025hunyuan3d}. Trained on large-scale curated subsets of 3D datasets such as Objaverse-XL~\cite{deitke23objaverse-xl:}, these models can generate a large variety of different objects.

However, to the best of our knowledge, no prior work has leveraged object generators for scene generation.

\section{Method}%
\label{sec:method}

Our goal is to generate a 3D world $\mathcal{G}$ from an initial textual prompt $p_0$.
Our main result is to show how prompt engineering can be used in combination with \emph{off-the-shelf} language, 2D and 3D generators to create the entire world automatically, with no need to retrain the models.

\begin{figure}[h]
\centering
\includegraphics[width=0.45\textwidth]{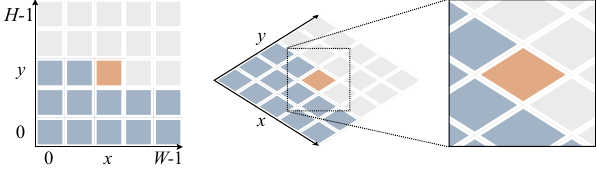}
\caption{\emph{Left:} Progressive generation of world tiles $\mathcal{T}$.
\emph{Right:} Isometric framing of a tile for image-based prompting.
}\label{fig:schema}
\end{figure}

We structure the world as a $W\times H$ grid $\mathcal{T} = \{0, \ldots, W-1\} \times \{0, \ldots, H-1\}$ of square tiles, each of which can contain several complex 3D objects (e.g., a building, a bridge, trees, etc.) as well as the ground surface.
We generate the world progressively, tile by tile, 
as shown in in \cref{fig:schema}.
Hence, when tile $(x,y)\in\mathcal{T}$ is generated, tiles 
$
\mathcal{T}(x,y)
=
\{(x',y')\in\mathcal{T} : y'<y \vee (y'=y \wedge x' < x)\}
$
have already been generated.

An overview of our approach is shown in \cref{fig:main}.
The first step of our method is to expand the world description $p_0$ into tile-specific prompts (\cref{sec:language-prompting}).
The second step is to pass these tile-specific prompts to a 2D image generator and inpainter to create an isometric view of each tile, accounting for the part of the world generated so far (\cref{sec:2d-prompting}).
The third step is to extract image prompts from these isometric views and use them as input to a image-to-3D generator to reconstruct each tile's geometry and appearance in 3D (\cref{sec:3d-prompting}).
The final step is to align and blend the 3D reconstructions of the tiles to create a coherent 3D world (\cref{sec:3d-fusion}).

\subsection{Prompting the Language Model}%
\label{sec:language-prompting}

The goal of language prompting is to take a high-level textual description of the world $p_0$ and expand it into a set $p$ of tile-specific textual prompts that can be used to generate the 3D world.
Specifically, $p$ is a collection of sub-prompts $p_{xy} \in \Sigma^\ast$, one for each tile, and a world-level `style' prompt $p_\star \in \Sigma^\ast$, so that we can write $p = \{p_{xy}\}_{(x,y)\in \mathcal{T}} \cup \{ p_\star \}$, where $\Sigma^*$ is the set of all possible strings.

The prompt $p$ could be constructed manually (which allows controlling the content of each tile) or generated by a large language model (LLM) such as ChatGPT~\cite{openai24gpt-4} from a `seed' prompt.
For the latter, we prompt ChatGPT o3-mini-high to generate a grid-like world with tile-specific descriptions, providing it with an example JSON file (see \cref{sec:lm-prompting-details} for details).

\subsection{Prompting the 2D Generator}%
\label{sec:2d-prompting}

\begin{figure}[h]
\centering
\includegraphics[width=\linewidth]{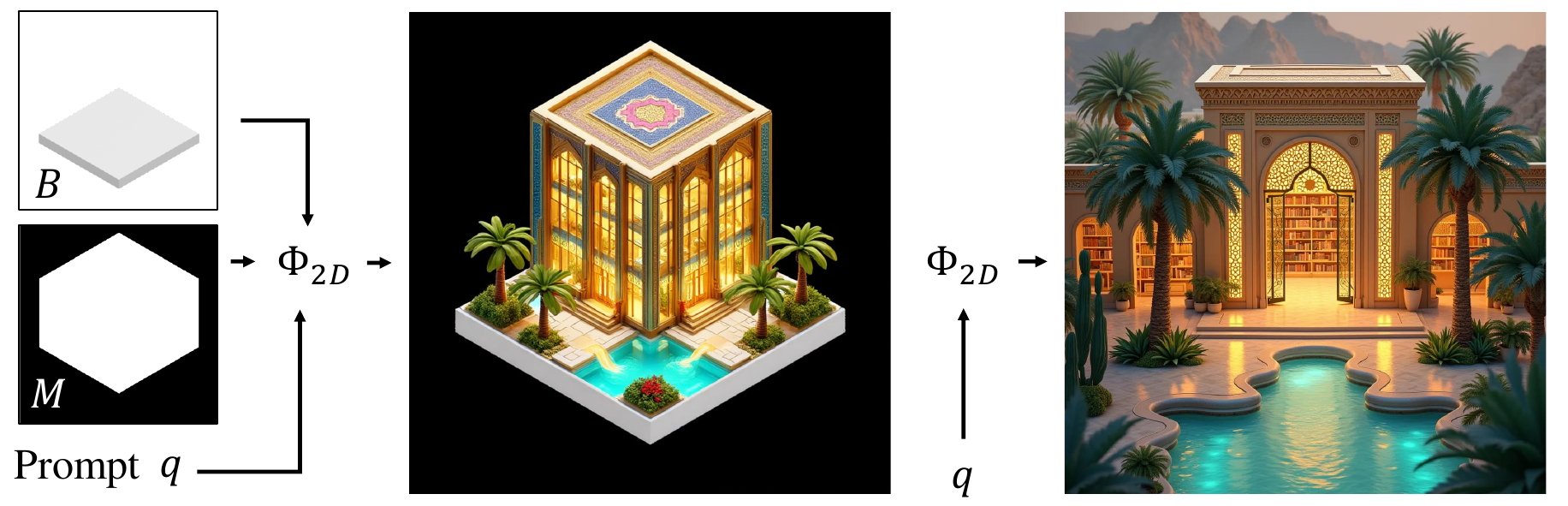}%
\caption{%
\emph{Left:}
Generation of the 2D image prompt for the first world tile at $x=0$ and $y=0$.
The image generator $\Phi_\text{2D}$ is conditioned on $q = p_{00} \cdot p_\star$ and tasked with inpainting the base image $B$ in the masked region $M$.
\emph{Right:}
If we do not `frame' the image by using $B$ and $M$, the generator produces an image which is not suitable for tiling.
}\label{fig:2d-prompting}
\end{figure}

We use the language prompts $p$ from \cref{sec:language-prompting} to prompt an off-the-shelf 2D image generator $\Phi_\text{2D}$ to output a 2D image $I(x,y)$ of each tile to be generated, as shown for example in \cref{fig:2d-prompting}. 
The image $I(x,y)$ must satisfy several constraints:
(1) It must reflect the tile-specific instructions $p_{xy}$ of the target tile as well as the world-level instructions $p_\star$.
(2) It must be suitable for prompting the image-to-3D generator in the next step.
(3) It must be consistent with the previously generated tiles.

Our prompting strategy is designed to satisfy these constraints.
The image is drawn as a sample $I(x,y) \sim \Phi_\text{2D}(q, B, M)$ from the 2D image generator $\Phi_\text{2D}$, where $q = p_{xy} \cdot p_\star$ is a prompt that combines the tile-specific and world-level descriptions.
The generator $\Phi_\text{2D}$ is also provided with a base image $B$ and an inpainting mask $M$ that constrain the output.
We assume that $\Phi_\text{2D}$ is capable of inpainting\,---\,a common feature of modern image generators.

\paragraph{Tile inpainting.}

To satisfy constraint (2), we need to encourage the image generator to generate regular tiles so that the image-to-3D model can output tiles with regular geometry that fit well together.
We assume that tiles have a fixed square basis of unit size and that they are imaged in an `isometric' manner.
This framing of the tiles is conducive to the generation of regular 3D tiles.
Furthermore, it is a common choice in video games and might have been observed by the image generator during training as these are often trained on game-like data.

Hence, our goal is to condition the image generator $\Phi_\text{2D}$ to produce images of this kind.
While one possible approach is to fine-tune the generator on such images, we demonstrate that it is possible to obtain this effect through prompt engineering alone, avoiding any retraining. 
We achieve this by carefully constructing the inputs $B$ and $M$ as shown in~\cref{fig:2d-prompting}.
Specifically, we set $B$ to be the image of the base of the tile, as a square, grey slab imaged from a fixed isometric vantage point.
The mask $M$ is a binary mask covering a cube on top of the base.

\Cref{fig:2d-prompting} shows the result of prompting the model in this manner as well as what happens if signals $B$ and $M$ are removed: the viewpoint and general frame of the tile is random and not suitable for 3D generation.

\paragraph{Taking the context into account.} 

\begin{figure}[ht]
\centering
\includegraphics[width=\linewidth,clip,trim=0pt 0pt 0pt 10cm]{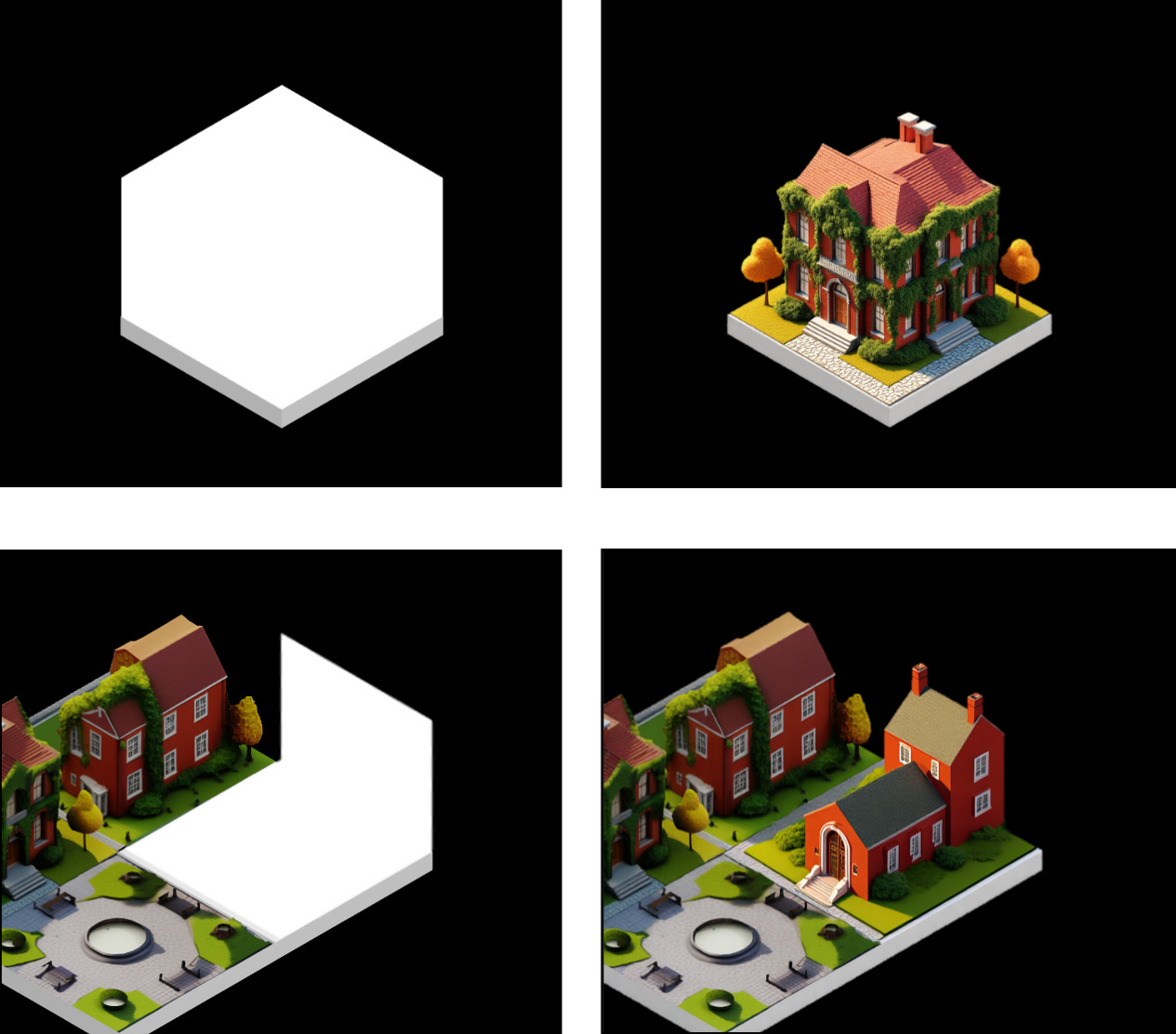}
\caption{
\emph{Left:}
Base image $B$ and inpainting mask $M$ (white overlay) to prompt the image generator $\Phi_\text{2D}$ to generate an image for a $x > 0$, $y> 0$ world tile.
\emph{Right:}
Result of inpainting.
}%
\label{fig:mask}
\end{figure}

Except for the first tile $(0,0)$, the tiles are generated in the context of the world already generated.
In order to account for this context, for tiles with $x, y >0$, we modify the base image $B$ and the mask $M$ as shown in~\cref{fig:mask}.
For the base image $B$, we render the part of the 3D world generated so far, providing context for the inpainting network.
We also modify the mask $M$ to avoid covering tiles already generated to the left (`west'; \ie, for a tile $(i, j) \in \mathcal{T}$ these are tiles $\{(x, y) \in \mathcal{T} : x < i \land y = j\}$).

\begin{figure}[ht]
\centering
\includegraphics[width=0.35\textwidth]{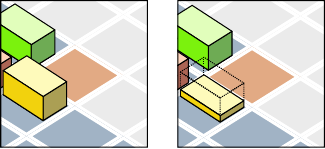}
\caption{Trimming tall structures for 2D prompting.
}
\label{fig:chop}
\end{figure}

Because we wish to ensure continuity of the ground, before rendering this contextual image, we trim any 3D geometry that is sufficiently high to occlude the tile we wish to generate, as shown in \cref{fig:chop} (see the result in~\cref{fig:mask}).

The appendix discusses a special case for tiles at the boundaries of the world (see \cref{sec:2d-prompting-details}).

\subsection{Prompting the 3D Generator}%
\label{sec:3d-prompting}

Given the tile image $I(x,y)$ obtained from the 2D image generator in \cref{sec:2d-prompting}, the next goal is to generate a corresponding 3D reconstruction  $G(x,y)$ of the tile utilizing an image-to-3D model $\Phi_\text{3D}$.
We opt for using a robust 3D generator and select TRELLIS~\cite{xiang24structured} due to its good performance, ability to generate both shape and texture, and latent space structure, which will be easy to manipulate for blending as we show later in \cref{sec:3d-fusion}.

Hence, 3D reconstruction amounts to drawing a sample $G(x,y) \sim \Phi_\text{3D}(J(x,y))$ from the image-to-3D generator $\Phi_\text{3D}$.
Rather than conditioning on the image $I(x,y)$, we use a pre-processed version $J(x,y)$, as described next.

\paragraph{2D foreground extraction and rebasing.}

\begin{figure}[ht]
\centering
\includegraphics[width=\linewidth,trim={0 22cm 0 0},clip]{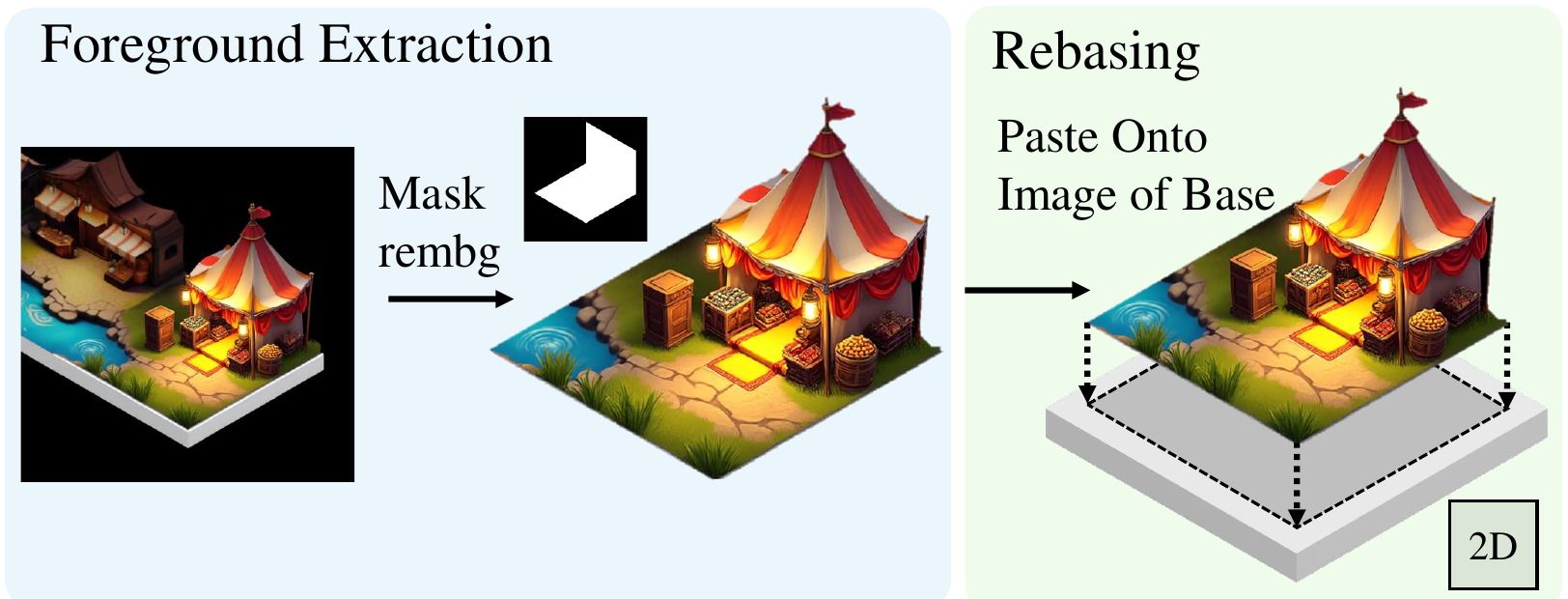}
\caption{\emph{Left:} Isolating the image of the new tile from $I(x,y)$.
\emph{Right:} Placing a slightly larger base underneath.
}%
\label{fig:rebasing}
\end{figure}

Recall that the image $I(x,y)$ output by the 2D generator of \cref{sec:2d-prompting} is an image of the tile \textit{and} its context.
However, the 3D generator $\Phi_\text{3D}$ expects the input image to only show the object that needs to be reconstructed, \ie, the new tile.
The first step is thus extracting only that part from $I(x,y)$ that corresponds to the new tile, which we do by applying the inpainting mask and then running \texttt{rembg}~\cite{Gatis_rembg_2025} with alpha matting~\cite{brinkmann2008art} to remove the background, as shown in \cref{fig:rebasing}.

The resulting image is narrowly cropped around the new tile.
Similar to \cref{sec:2d-prompting}, we found it beneficial to hallucinate a base for the tile, an operation that we call `rebasing', as shown in \cref{fig:rebasing}.
We simply compose the image of the tile with a slightly larger gray slab (in 2D) to obtain $J(x,y)$, which in effect provides a `frame' for the 3D generator to work with.
The base is reconstructed as part of the tile's geometry, which can be used for validation and as a simple-to-detect handle for further 3D processing.

\begin{figure}[ht]
\centering
\includegraphics[width=0.48\textwidth]
{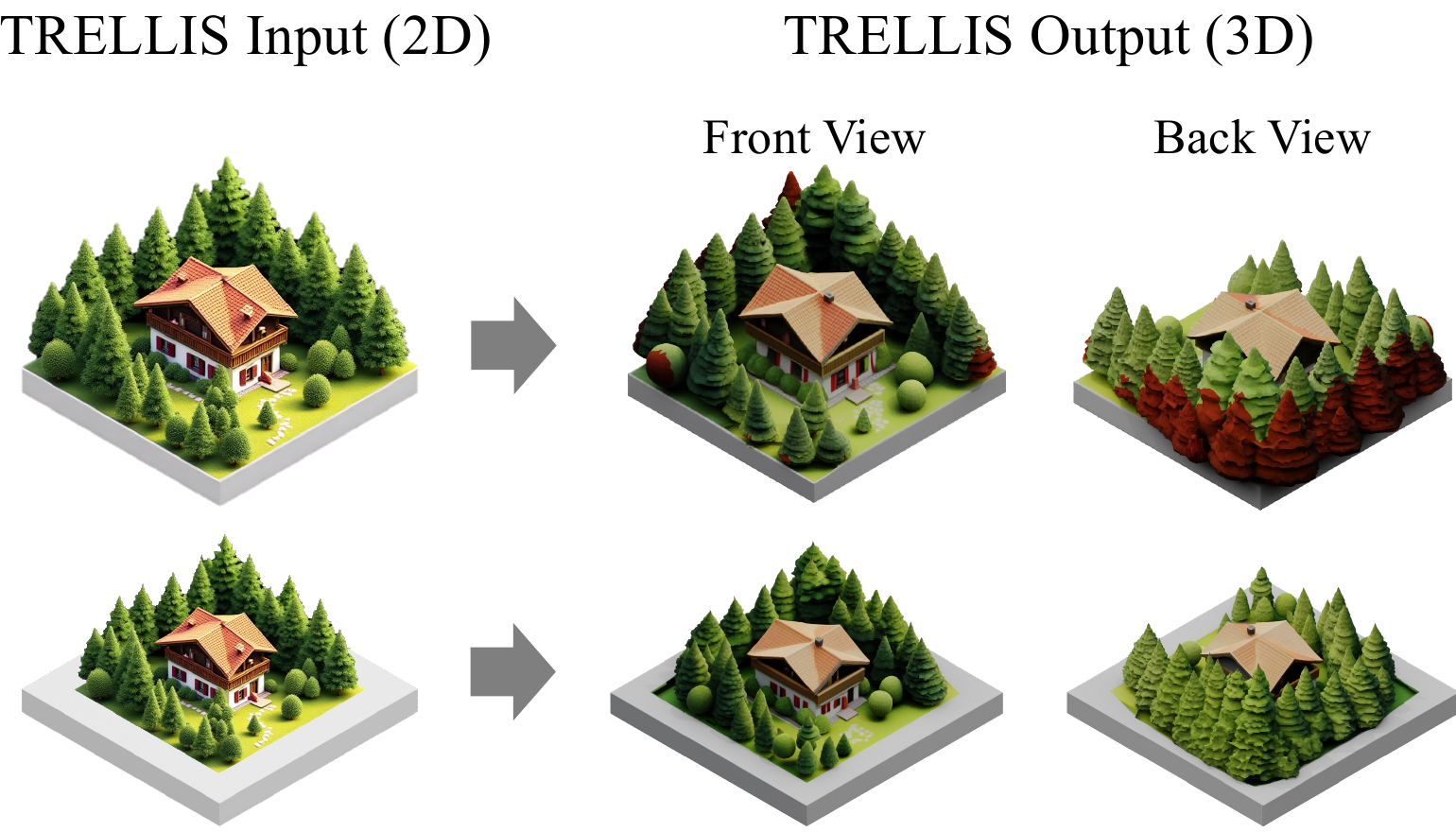}
\caption{
\emph{Top:} 3D reconstruction using a tight base.
\emph{Bottom:} the same, but with a slightly larger base, which helps to contain the tile's geometry above ground (see the back of the reconstruction), and creates an easy-to-detect 3D base.
}%
\label{fig:rebasing-abl}
\end{figure}

The `rebased' image $J(x,y)$ is fed to the 3D generator $\Phi_\text{3D}$ to obtain the 3D reconstruction $G(x,y)$ of the tile, which are 3D Gaussian Splats~(3DGS).
The effect of rebasing on the 3D result is shown in \cref{fig:rebasing-abl}.

\paragraph{3D geometric validation.}

Because the generators are imperfect, we verify the 3D reconstruction $G(x,y)$ to ensure that it is of sufficient quality.
If not, we discard it and generate the tile again using a different random seed.
To verify the tile, we use a few heuristics that check that the tile's geometry occupies a square region of sufficient size and that the base of the tile has been reconstructed faithfully.
Please see \cref{sec:3d-geometry-validation-details}.

\paragraph{3D post-processing.}

At this point, we have verified the 3D reconstruction $G(x,y)$ for the tile as a mixture of 3D Gaussians.
However, the actual 3D footprint, orientation, and size of the tile are controlled by the 3D generator and are inconsistent.
The post-processing step applies simple heuristics to refine the 3D Gaussian representation by first cropping out the added base, then rescaling the tile to a unit size, and finally reorienting it to match the 2D image prompt.
We explain this in more detail in \cref{sec:post-processing-details}.

\subsection{3D Blending}%
\label{sec:3d-fusion}

At this point of the pipeline, we have reconstructed all 3D tiles $G(x,y)$, $(x,y)\in\mathcal{T}$.
As a result of the prompting and post-processing steps in \cref{sec:2d-prompting,sec:3d-prompting}, the tiles are already approximately aligned and correctly oriented, with their ground level at roughly the same height in 3D space.
Because the 3D reconstructions are 3D Gaussian Splats, it is easy to simply take their union as the reconstruction of the whole world.

Even so, the boundaries of the tiles may not match perfectly.
This is largely due to the fact that TRELLIS does not reconstruct the input images exactly and to the fact that only a single view of each tile is provided to it, which only indirectly controls the reconstruction of the back of the tile.
In this section, we thus propose a method to improve the blending of tiles, ensuring that the 3D world is coherent and continuous.

In particular, we regenerate the boundary region of two adjacent tiles in the latent space of TRELLIS, in essence \textit{blending} it, and then decode it into 3DGS\@.
Next, we discuss the specifics of this process.

\paragraph{Blending in 2D.} 

To blend the latents of two neighbouring tiles, we first predict the appearance of the boundary between the two tiles. 
To that end, we place the two 3D tiles next to each other, render a frontal view, and inpaint the middle region of the rendering (\cref{fig:main}) with $\Phi_\text{2D}$. 
This leads to a well-blended image, which we use to condition for $\Phi_\text{3D}$.

\paragraph{Blending in 3D.}

Next, we use $\Phi_\text{3D}$ to blend the latents. 
We take the latents of the two tiles $\gamma^1$ and $\gamma^2$ where $\gamma^1, \gamma^2 \in \mathbb{R}^{D \times R \times R \times R}$ are $D$-dimensional features in the $R$-sized 3D grid that TRELLIS denoises. We put them together in a new volume $\gamma$, where the side where they meet is in the middle:
$$
\gamma_{:, x, y, z} =
\begin{cases}
\gamma^{1}_{:, x + R/2, y, z}, & \text{if } x < R/2 \\
\gamma^{2}_{:, x - R/2, y, z}, & \text{if } x \geq R/2.
\end{cases}
$$

We apply the denoising function $\Omega$, which is the latent denoiser of $\Phi_\text{3D}$, to the volume $\gamma$, but only within the middle region where we have applied the stitch, \ie, for $x \in [R/2 - r, R/2 + r]$ for some $r < R/2$, while keeping the rest fixed.
Formally, we initialise $\tilde{\gamma} \sim \mathcal{N}(0, I)$ and at each denoising step $t$, we update $\tilde{\gamma}$ as:
$$
\tilde{\gamma}_{t+1, :, x, y, z} =
\begin{cases}
\Omega(\tilde{\gamma}_{t,:, x, y, z}), & \text{if } |x - R/2| \leq r \\
\gamma_{t+1, :, x, y, z}, & \text{otherwise},
\end{cases}
$$
where $\gamma_t$ is obtained by adding noise to the original $\gamma$ at the corresponding noise level for step $t$. 
In practice, we only denoise the second stage of TRELLIS, keeping the occupancy latents of the first stage fixed. 
The reason for that is that the first stage is at a very low spatial resolution ($R=16$, compared to $R=64$ of the second stage), which gives little flexibility for the size of the denoised region $r$.

\begin{figure}[t]
\centering
\includegraphics[width=0.99\linewidth,trim={3.5cm 8.5cm 10.5cm 4cm},clip]{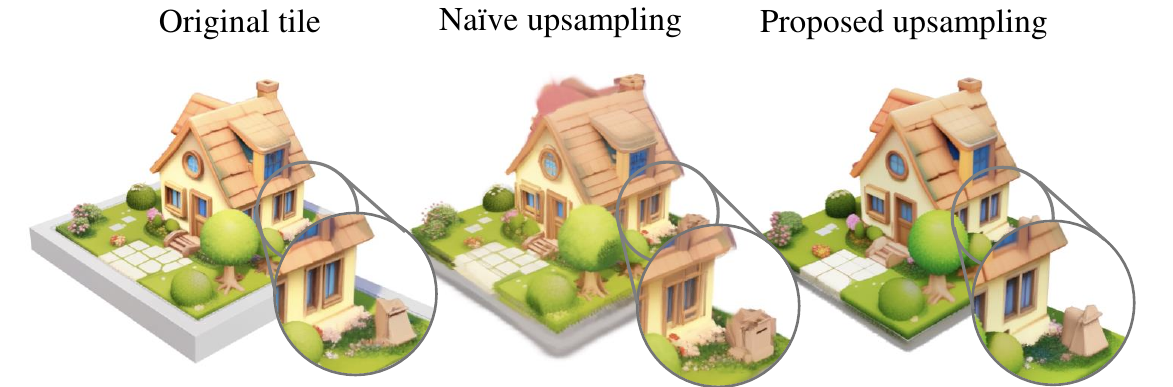}
\caption{\textbf{Upsampling sparse latents.} We need to resize or upsample sparse latents in order to stitch them. Due to the sparsity of the latents and the behaviour of the latent decoder, naively resampling in latent space leads to artifacts. Our proposed resizing of the sparse latents better preserves textures and fine structures.}%
\label{fig:upsampling}
\end{figure}

\paragraph{Upsampling the latents.}

Remember that due to the rebasing, $G(x,y)$ contains a 3D base. While we have removed the base in 3DGS space, we have yet to do the same in the latent space. We use the same cuts we applied in 3DGS space to now crop the latents, rounding the cuts to account for the discrete nature of the latent voxel grid. In the previous step, however, we assumed that the latents $\gamma^1$ and $\gamma^2$ have the same spatial resolution, $\gamma \in \mathbb{R}^{D \times R \times R \times R}$. After cropping, this is not the case any more if the cuts of neighbouring tiles differ.
Thus, similarly to how we resize the tiles' 3DGS to a unit size, we have to upsample the now-cropped latents back to the original grid resolution $R$.

We found that naively upsampling the latents by interpolation leads to poor reconstructions, as shown in~\cref{fig:upsampling}.
We attribute this to the sparse structure of the latents and quirks of the latent decoder of TRELLIS. 

We propose the following upsampling scheme. 
First, we upsample the cropped occupancy volume that TRELLIS predicted to the original resolution $V \in \{0, 1\}^{R\times R \times R}$. 
Next, we denoise \emph{a new set} of latents $\gamma$ on the upsampled occupancy volume. 
To preserve the details and textures of the original 3D tile, we render it from multiple views and jointly condition the denoising on all of them.
In practice, when denoising with multiple conditioning views, at each timestep, the denoising step is computed as the average denoising step across all views.
We show that this upsampling scheme leads to superior reconstructions in~\cref{fig:upsampling}.

\section{Experiments}%
\label{sec:experiments}

\begin{table}[t]
    \centering
    \small
    \begin{tabular}{cccccc}
        \toprule
        \multicolumn{5}{c}{\textbf{Win Rate (\%)}} \\
        Overall & Geometry & Exploration & Diversity & Realism \\
        \midrule
        90.9 & 81.8 & 90.9 & 90.9 & 86.4 \\
        \bottomrule
    \end{tabular}
    \caption{Win rates of our method against BlockFusion. We asked participants to select which scene they prefer overall, as well as which one has better geometry, would be more interesting to explore, is more diverse, and has better realism.}
    \label{tab:win_rates}
\end{table}
\begin{figure}[t]
\centering
\includegraphics[width=0.48\textwidth]{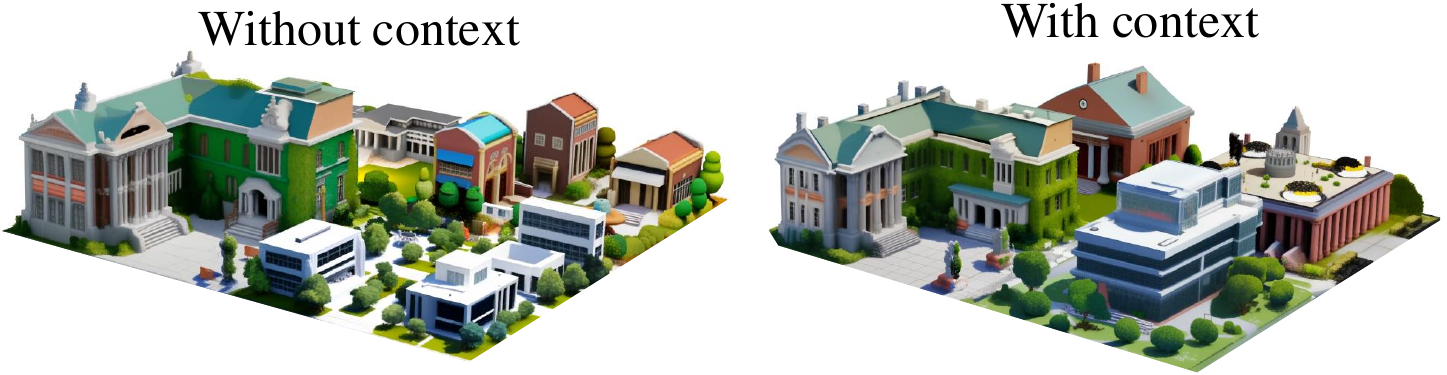}
\caption{\textit{Left:} 2$\times$2 grid generated with our method, where context is taken into account as described in~\cref{sec:2d-prompting} \textit{Right:} Generated with our method using the same prompts, but \emph{not} taking into account context --- here, the scale of the buildings is not consistent.}%
\label{fig:context}
\end{figure}
\begin{figure*}[t]
\centering
\includegraphics[width=\textwidth]{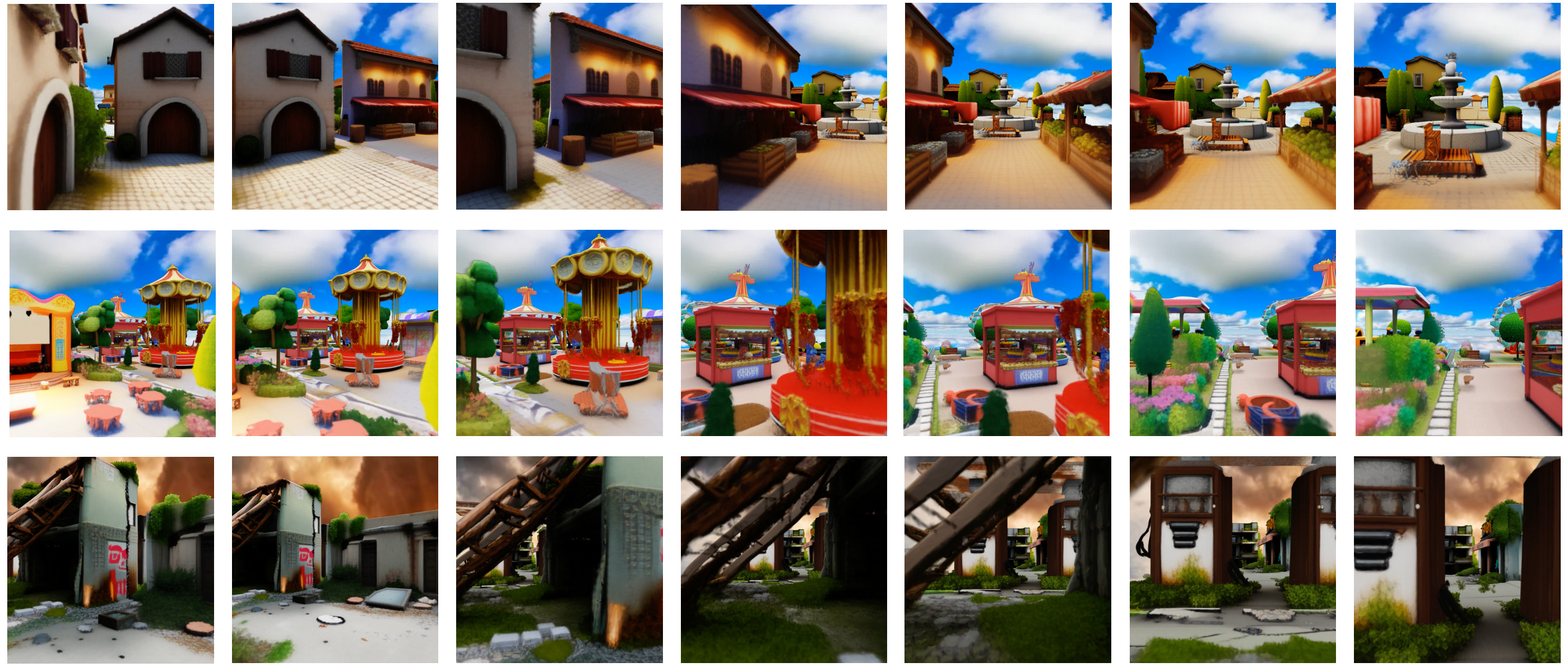}
\caption{\textbf{Exploring a 3D world.} We show trajectories exploring the 3D worlds we generate. Please see the supplement for videos.}
\label{fig:explore}
\end{figure*}

\paragraph{Experimental details}

We generate the text prompts using ChatGPT o3-mini-high.
For the 2D inpainter, we use the Flux ControlNet of~\cite{alimama2024fluxcontrolnet}.

\paragraph{Human preference.}

We evaluate human preference for the results generated by our method and those obtained with BlockFusion~\cite{wu24blockfusion:}.
In particular, we compare a `city' scene, showing the entire scene as well as close-up detail views.
As seen in~\cref{tab:win_rates}, participants ($n=22$) find our method better overall, with better geometry, realism, and diversity.

\subsection{Ablations}

Here we ablate several components of our approach, showing the importance of each of them.

\paragraph{Building a grid.}

A naive approach to generating a 3D scene is querying the image generator to produce an image of a large-scale scene (using our 2D image prompt setup) and then obtaining the entire 3D world directly with TRELLIS\@.
To achieve the same level of control provided by our method, the textual prompt needs to be highly detailed and include layout instructions.
However, we found neither precise nor abstract prompts to be effective at steering the generations of Flux (for details, see A.4).

\paragraph{2D prompting context.}

We remove context from neighbouring tiles, as described in~\cref{sec:2d-prompting}.
Doing that, each tile is sampled independently, and the relative scale between objects is inconsistent, as shown in~\cref{fig:context}.

\paragraph{Rebasing.}

To place tiles on a grid, they need to be square (otherwise the grid would be jagged) and their base needs to have been reconstructed faithfully (clearly delimiting where the tile stops).
Without rebasing, the geometry generated by TRELLIS might extend beyond the base and makes the tile's `true' extent difficult to detect, as shown in~\cref{fig:rebasing-abl}.
We ablate the effect of rebasing, using a small $2 \times 2$ scene to curb the effect of error accumulation.
As seen in~\cref{tab:rebasing}, no rebasing causes TRELLIS to generate tiles that are, on average, neither perfectly square nor have a solid border.

\begin{table}[]
\small
\begin{tabular}{lccc}
\toprule
Method       & Base Area & Squareness $\uparrow$ & Completeness $\uparrow$ \\
\midrule
No Rebasing  & 2271      & 0.92                   & 0.73   \\
Ours         & 4096      & \textbf{1.00}          & \textbf{1.00} \\
\bottomrule
\end{tabular}
\caption{Average tile 3D geometry metrics for an approach without rebasing and our method. Rebasing is crucial to ensure a tile is square and its base has been reconstructed faithfully.
The metrics we use are the area of the base in voxels, a measure for the `squareness' of the base, and how many border voxels have been faithfully reconstructed. For details, please refer to the appendix.}
\label{tab:rebasing}
\end{table}
\begin{table}[ht]
\centering
\small
\resizebox{0.45\textwidth}{!}{%
\begin{tabular}{lcccc}
\toprule
Method & LPIPS $\downarrow$ & SSIM $\uparrow$ & FID $\downarrow$ & KID $\downarrow$ \\
\midrule
Naive  upsampling               & 0.5914         & 0.3093         & 200.5         & 0.243         \\
Ours (single frame)   & 0.3517         & 0.5149         & 111.6         & 0.069         \\
Ours (multi frame)    & \textbf{0.3212} & \textbf{0.5312} & \textbf{89.1} & \textbf{0.051} \\
\bottomrule
\end{tabular}%
}
\caption{Perceptual metrics for our methods and the naive approach. Lower values for LPIPS~\cite{zhang18the-unreasonable}, FID~\cite{heusel17gans}, and KID~\cite{binkowski2018demystifying} are better, while higher values for SSIM are better. We see that even using a single conditioning frame leads to better upsampling results, and multiple frames further improve performance.}
\label{tab:perceptual}
\end{table}

\paragraph{Latent upsampling.}

We sample 10 random views each from 200 tiles generated by TRELLIS and compute perceptual metrics in~\cref{tab:perceptual} when upsampling with our proposed upsampling approach in~\cref{sec:3d-fusion} and a naive interpolation.
We see that the proposed method leads to better results across a range of metrics, even when using a single conditioning frame.

\paragraph{3D blending.}

\begin{figure}[t]
\centering
\includegraphics[width=\linewidth]{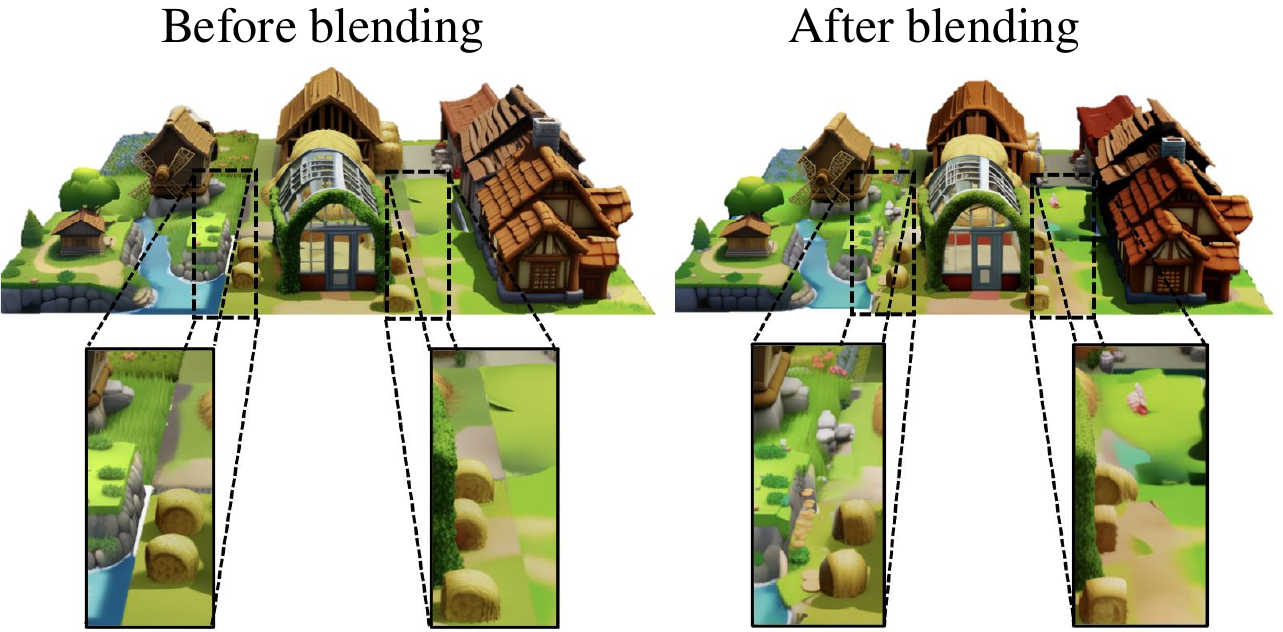}%
\caption{%
\emph{Left:}
Tiles before applying the 3D blending step
\emph{Right:}
After the 3D blending step. We see that where boundaries between tiles were obvious, they are now well-blended.
}\label{fig:blending}
\vspace{2em}
\end{figure}
In~\cref{fig:blending}, we generate a scene where we do not apply the 3D blending (\cref{sec:3d-fusion}), resulting in discontinuities between the tiles.

\subsection{Qualitative Results}

We present example scenes generated by our method in~\cref{fig:splash}.
Further, we show detail views, highlighting the quality and diversity of the scene.
Please see the supplementary material for many more examples.

\paragraph{Exploring a generated world.}

We can sample trajectories exploring the generated 3D worlds (\cref{fig:explore}).
A skybox has been added for visual effect.
Unlike the trajectories generated by world video models~\cite{parker-holder24genie}, ours are guaranteed to be consistent and do not suffer from semantic drift.
Different from other systems that only generate a bubble, our method creates spaces sufficiently large to be navigated in a non-trivial manner.

\section{Conclusion}%
\label{sec:conclusions}

We have introduced \method, an approach to generate diverse, high-quality, and complex 3D worlds with fine-grained control over their layout and appearance.
\method creates worlds by autoregressively generating tiles on a grid, enabling scalability to arbitrary grid sizes. 
By accounting for local context and by means of 3D inpainting, the tiles are seamlessly stitched together into coherent scenes.
\method is flexible: it can 
either generate worlds from a brief `world' text prompt
or allow control of the individual tiles via tile-specific instructions, all the while maintaining an overall thematic consistency of the generated world.
The rich detail of the generated worlds can be fully explored, not restricted to a single `3D bubble' as in many prior works.

We have demonstrated the effectiveness of off-the-shelf generation by utilizing pre-trained language, 2D, and 3D generators through carefully designed prompting strategies and without requiring retraining of any of these components.
Nevertheless, we expect that in cases where 3D scene scale data is available, fine-tuning some components would result in further improved results and simplifications in the alignment and rebasing steps of the pipeline.
Future work could also consider relaxing the tile structure, for example, by randomly shifting and scaling tiles and using coarse-to-fine modeling to ensure coherent global structure and fine-grained local details.

\paragraph*{Ethics.}
For details on ethics, data protection, and copyright, please see \url{https://www.robots.ox.ac.uk/~vedaldi/research/union/ethics.html}.

\paragraph*{Acknowledgments.}

The authors of this work are supported by ERC 101001212-UNION, AIMS EP/S024050/1, and Meta Research.

{
\small
\bibliographystyle{ieeenat_fullname}
\bibliography{vedaldi_general,vedaldi_specific,local}
}

\cleardoublepage\newpage
\maketitlesupplementary
\appendix
\section{Appendix}

\subsection{Language Model Prompting Details}%
\label{sec:lm-prompting-details}

While the prompt $p$ can be constructed manually, an LLM may also be employed. For it to understand the task it is asked to solve, we utilize the following prompt for ChatGPT o3-mini-high:

\begin{quotation}
\noindent
Assume you had access to an AI model that can generate small-scale cities on an isometric grid by creating individual tiles. For each of these tiles (identified by their 2D position), a short but expressive text prompt has to be provided. Additionally, a global prompt is used, which provides context, lighting, time of day, as well as the art style. The prompts of the tiles can be generic but they might have a semantic connection to neighbouring tiles (such that a river can flow through the city on multiple tiles). The format for the instructions to the AI model is JSON.
Consider the following example: \texttt{<TEMPLATE>}
The art style and perspective mentioned in the prompt should be maintained. The rest may be freely adapted. There are no limits to the setting, the sky is your limit. Now, please generate a $3 \times 3$ grid.
\end{quotation}

In this prompt, a template or a `seed' prompt is provided. We use a simple JSON file, as exemplified in \cref{fig:lang-prompt}.
% chktex-file 18
\begin{figure}[h]
\begin{roundedlisting}
{"tiles": [
{ "prompt": "ancient stone bridge over a stream", "x": 0, "y": 0 },
{ "prompt": "lively stream past mossy banks", "x": 1, "y": 0 },
{ "prompt": "serene pond reflecting moonlight", "x": 0, "y": 1 },
{ "prompt": "bustling medieval market street", "x": 1, "y": 1 } ],
"prompt": "{tile_prompt}, medieval setting, isometric view, glowing lanterns, soft shading, vibrant colors, detailed textures"}
\end{roundedlisting}
\vspace{-1em}
\caption{Example JSON file to describe each tile in a $2\times 2$ world.}%
\label{fig:lang-prompt}
\end{figure}

\subsection{2D Prompting Details}
\label{sec:2d-prompting-details}
In Sec. 3.2, we describe how tiles are generated in the context of those that already exist. There is a special case that we address separately, where context has to be bootstrapped, namely tiles $\mathcal{L} :=\{(x, y) \in \mathcal{T} : x = 0 \land y > 0\}$. Due to our build order and the trimming of obstructing 3D geometry, these tiles might lack sufficient contextual cues. As a remedy, we temporarily provide context with a previously generated tile: For a tile $(0, y) \in \mathcal{L}$, we duplicate the tile $(0, y-1) \in \mathcal{T}$ and place the copy at position $(-1, y)$. During inpainting, this tile serves to provide context in terms of scale and general appearance. Once inpainting is completed, this copy is removed.

\definecolor{custom_orange}{RGB}{226,170,131}
\begin{figure}[h]
\centering
\includegraphics[width=\linewidth]{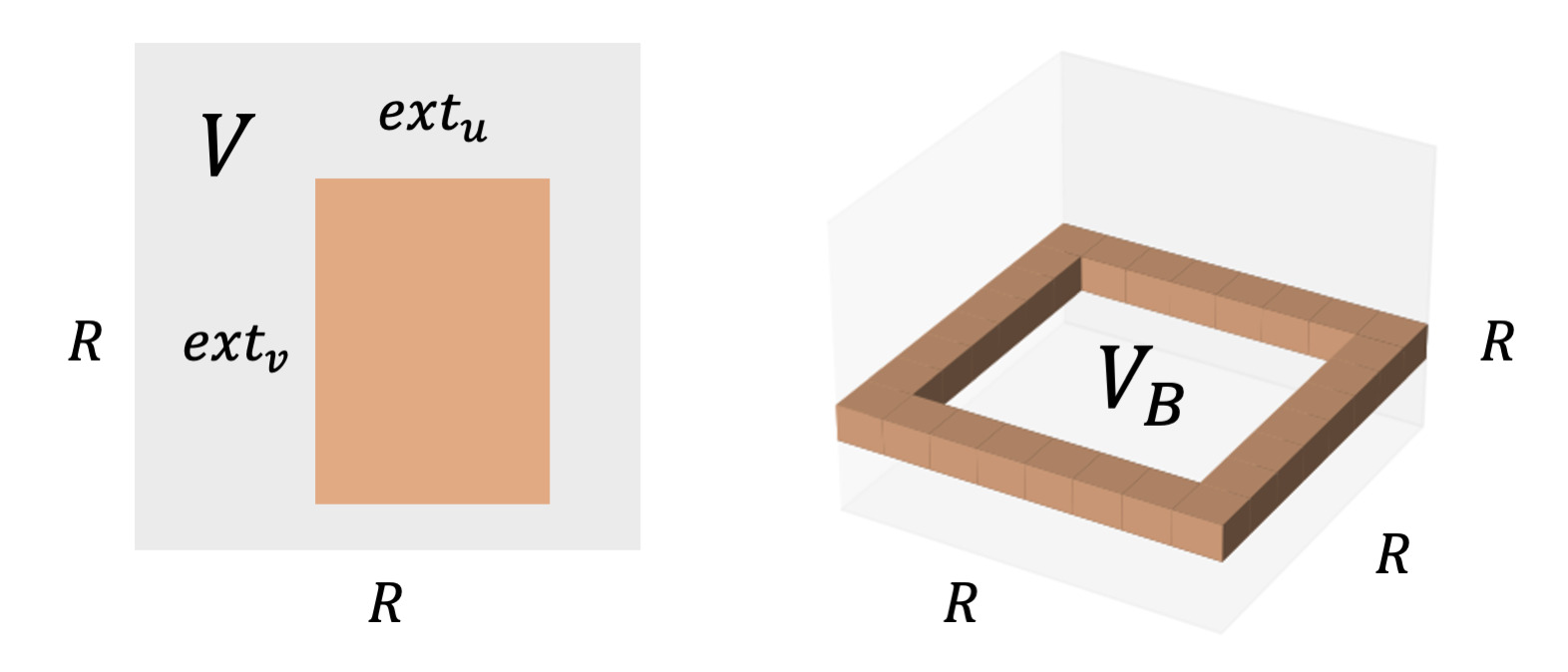}%
\caption{%
\textbf{Tile geometry validation.}
To check the geometric qualities of a reconstructed tile, we look at the occupancy grid $V \in \{0, 1\}^{R\times R \times R}$ generated by TRELLIS.
Activated voxels are indicated in orange (\textcolor{custom_orange}{$\blacksquare$}).
\emph{Left:}
The extent of an object in an object at height $w$ (slice visualized in 2D).
\emph{Right:}
An example of a 3D tile base template $V_B$.
}\label{fig:3d-verification}
\end{figure}

\subsection{3D Geometry Validation Details}%
\label{sec:3d-geometry-validation-details}

TRELLIS is a two-stage method and produces an occupancy volume $V \in \{0, 1\}^{R\times R \times R}$ in the first stage (before the 3D Gaussian mixture is output), where $R=64$ is the resolution of the grid.
To perform geometric validation,  we utilize this occupancy volume, which captures the rough 3D geometry of the tile, and check that it conforms to the desired geometry.

First, we test whether the reconstructed tile is supported by a square by computing its 2D rectangular footprint and ensuring that the latter is sufficiently large and isotropic.

To this end, let $(u,v,w)$ index the $R\times R \times R$ voxel grid, where $u$ and $v$ corresponds to world directions $x$ and $y$.
Let
$
(
    u_\text{min},
    u_\text{max},
    v_\text{min},
    v_\text{max},
)
$
be the bounding box containing all the active voxels
at height $w$.%
\footnote{
So for instance
$
u_\text{min}
=
\min \{u : \exists v,w: V(u,v,w) = 1\}.
$
}
Let 
$
\operatorname{ext}_u = 
\max
\{0, 1 + u_\text{max} - u_\text{min}\}
$
be the width of the bounding box and $\operatorname{ext}_v$ its height.
We discard the tile if the area is too small, \ie, if
$
\operatorname{ext}_u \cdot \operatorname{ext}_v < (R/2)^2
$.
We also discard it if it is not square, \ie, if
$
\min \{\operatorname{ext}_u, \operatorname{ext}_v\}
/
\max \{\operatorname{ext}_u, \operatorname{ext}_v\}
< \alpha = 1
$.

Second, we check that the base that we have added in 2D in the `rebasing' step has been faithfully reconstructed in 3D.

We define a 3D tile base template, which we call $V_B$. Let $u_\text{min}(w)$ be the minimum $u$ of the bounding box that contains the volume slice at height $w$, and define $u_\text{max}(w)$ and so on in a similar manner,
so for instance
$
u_\text{min}(w)
=
\min \{u : \exists v: V(u,v,w) = 1\}
$.
Let $w^\ast$ be the height at which the base is the largest, \ie,
$
w^\ast = \operatorname{argmax}_w \operatorname{ext}_u(w) \cdot \operatorname{ext}_v(w).
$
Then, $V_B$ is the indicator function of the voxels $(u,v,w)$ such that $w=w^*$ and
{\scriptsize
$$
\max
\left \{
\frac{
 |2 u - u_\text{max}(w^\ast) - u_\text{min}(w^\ast)|
}{
 u_\text{max}(w^\ast) - u_\text{min}(w^\ast)
},~
\frac{
 |2 v - v_\text{max}(w^\ast) - v_\text{min}(w^\ast)|
}{
 v_\text{max}(w^\ast) - v_\text{min}(w^\ast)
}
\right \}
= 1.
$$}%
Note that the template $V_B$ is constructed adaptively to match the input tile $V$.

We discard a generated tile if
$
(V \cdot V_B) / (V_B \cdot V_B) < \beta = 0.95 ,
$
where $\cdot$ denotes the inner product of tensors.

\subsection{3DGS Post-Processing Details}%
\label{sec:post-processing-details}

\paragraph{3D cropping, resizing and centering.}

Given the 3D Gaussian mixture $G(x,y)$ initially output by the 3D generator, we first identify the extent of the tile `proper' (discounting the extended base).
We consider the $xy$ footprint of the tile (\ie, we look at the tile from above) and seek to identify four cuts (from the left, right, top, and bottom) that define an axis-aligned rectangle strictly containing the tile.
For example, to determine the location of the left cut $x^*$, we consider slices
$V_x = \{(x',y',z') \in \mathbb{R}^3 : x-\delta \leq x' < x+\delta\}$.
We find the 3D Gaussians whose centers falls within $V_x$ and compute their average color $c_x$.
Then, we compute the distance $d(x) = \|c_x - c_{x_\text{min}}\|$ where $c_{x_\text{min}}$ is the average color of the leftmost slice (used as a reference).
We set $x^* = \min \{x : d(x) > \tau\}$ where $\tau$ is a threshold, which corresponds to the slice that transition from the `background' color to something else.

We find in this manner the four cuts, keep only the Gaussians contained in the resulting rectangular footprint, and recenter and resize this footprint to fill the standard tile size.

Additionally, the base allows us to figure out the position of the tile's surface:
As TRELLIS centers objects vertically, the ground surface level of any two tiles will vary.
We use the average height of the tile's four corners to determine the position of the surface, allowing us to align it with others.

\paragraph{3D reorientation.}

We also note that TRELLIS generates the 3D object with an arbitrary orientation with respect to the input image $\tilde I(x,y)$.
However, the tile must be inserted with the correct orientation in the 3D world a otherwise the continuity between tiles, which the inpainting method of \cref{fig:2d-prompting} encourages, will be lost.
In practice, the ambiguity is limited to 90-degree rotations around the vertical axis and is very easy to remove.\footnote{
This is likely due to the implicit bias in the TRELLIS training set that consists of synthetic 3D objects which are almost invariably axis aligned.
}
To do so, we test four possible 90-degree rotations of the tile around the vertical axis, rendering the corresponding views and comparing them to $\tilde I(x,y)$ using the LPIPS~\cite{zhang18the-unreasonable} loss.
The minimizer is taken as the correct orientation.

\subsection{Ablation Details}%
\label{sec:ablations-appendix}
\paragraph{Building a Grid.}
In the following, we present results from our experiments attempting to generate a large scene non-iteratively. Here, we generate single image with Flux that is used as conditioning for TRELLIS to directly create the desired scene.

\begin{figure}[ht]
\centering
\includegraphics[width=0.45\textwidth]{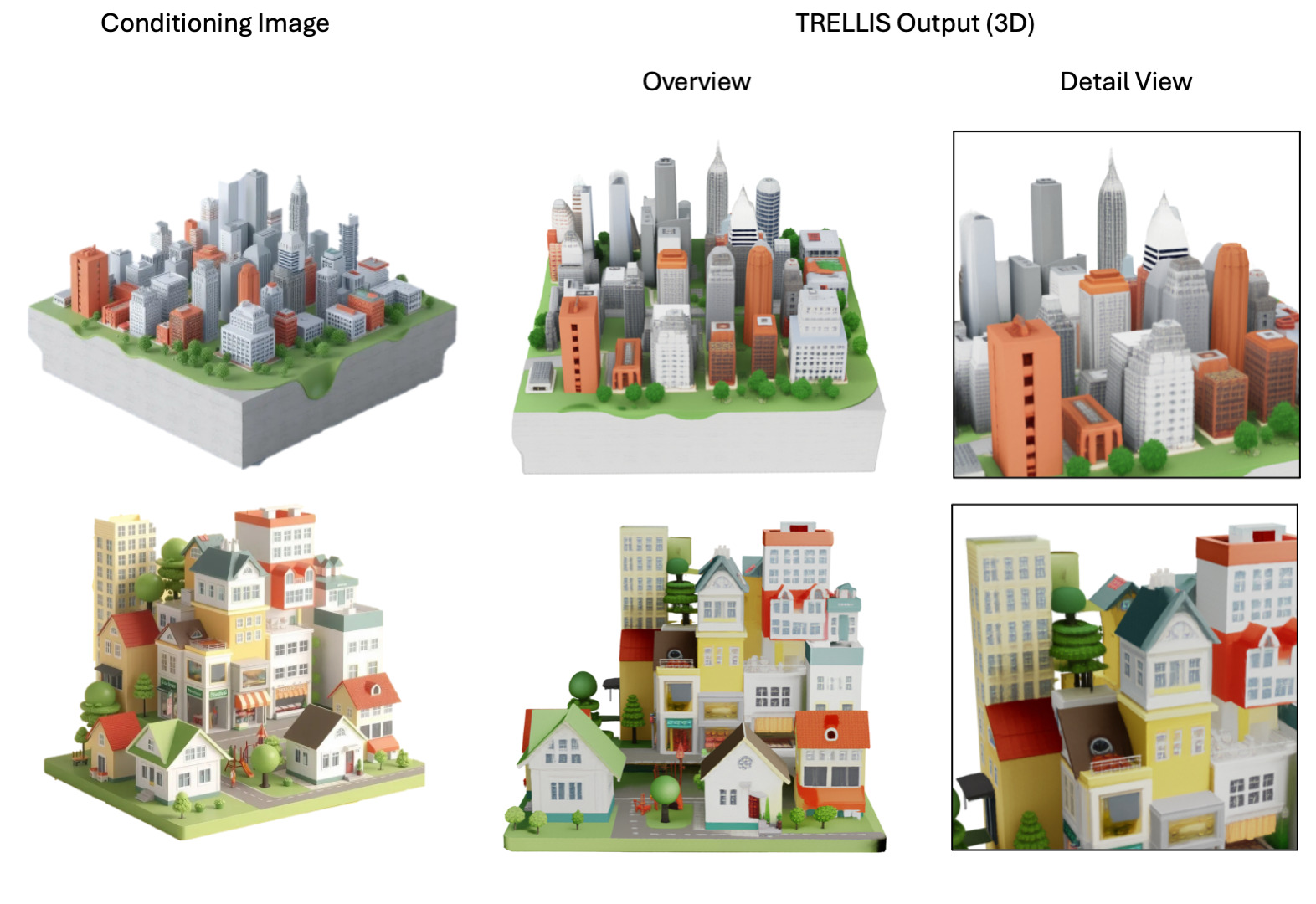}
\caption{\textbf{Non-iterative city building.} We obtain conditioning images generated by Flux (left) and directly use them to build a large-scale scene with TRELLIS (center). While the generated 3D structures are visually appealing, the level of detail (right) is very limited. The first row used generic prompting for the conditioning image (``a city scene on top of a base''), whereas the second row uses a more involved prompt with an explicit layout (e.g., ``a house in the bottom left corner'', ``a pharmacy in the top right corner'').}
\label{fig:city-prompting-naive}
\end{figure}

In the first set of experiments, we do not use our 2D prompting design. To obtain an isolated 3D object that can be generated by TRELLIS, we use prompts with the prefix ``a 3d object of''. We show those results in \cref{fig:city-prompting-naive}. While the generated objects are visually appealing, they have several limitations: (i) The resolution of the conditioning image and the 3D structures TRELLIS can generate is limited. Therefore, this approach is not scalable to arbitrarily large scenes. (ii) Due to the lack of perfect control over the base structure, the result cannot be easily extended or edited. (iii) The layout instructions are mostly ignored, thus severely limiting the level of control over the generation.

\begin{figure}[ht]
\centering
\includegraphics[width=0.45\textwidth]{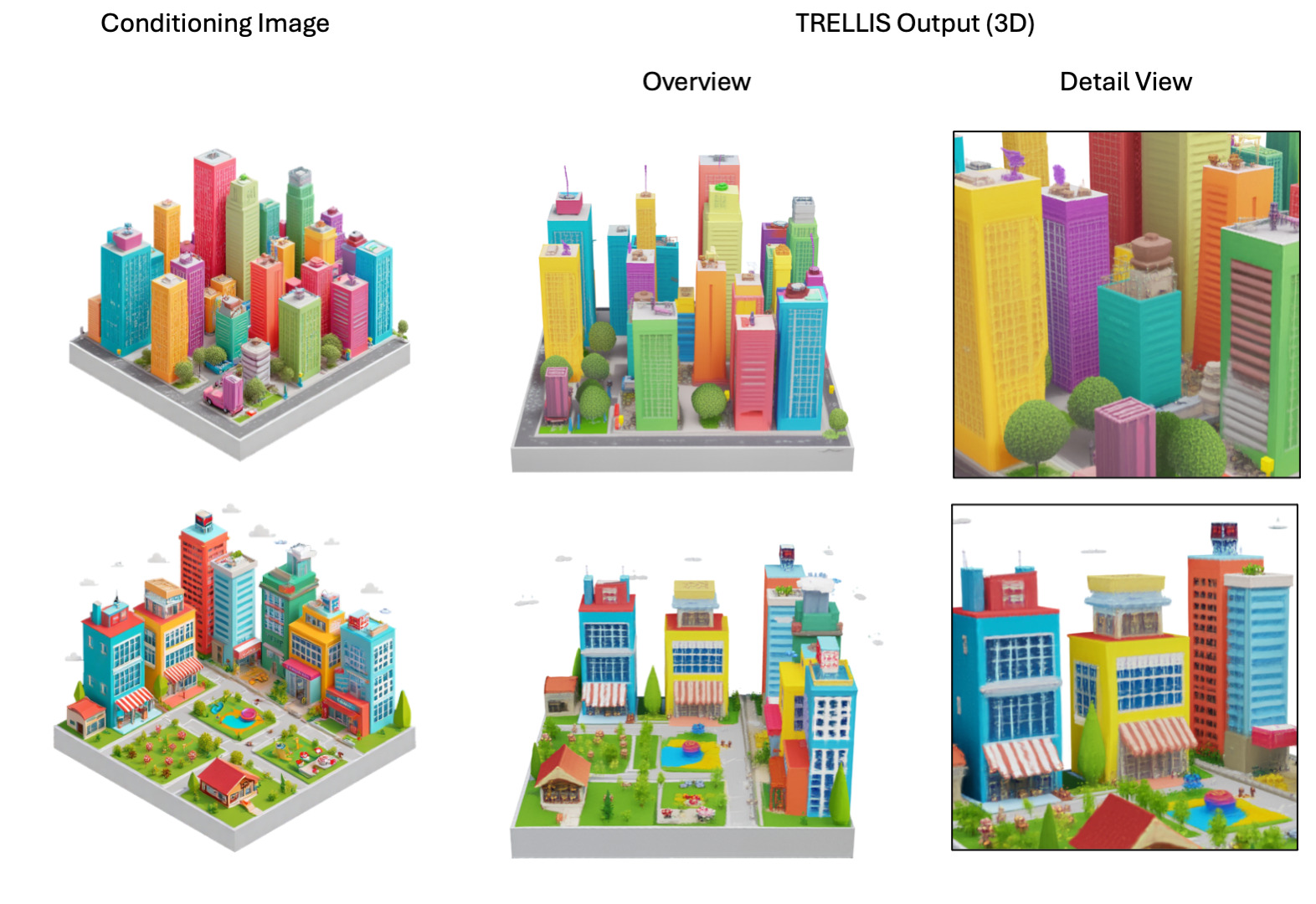}
\caption{\textbf{Non-iterative city building (with our 2D prompting).} We obtain conditioning images generated by Flux (left) and directly use them to build a large-scale scene with TRELLIS (center). Despite the initial visual appeal, the structures lack in detail. The first row used generic prompting for the conditioning image (``a vibrant city scene''), whereas the second row uses a more involved prompt with an explicit layout (e.g., ``a house in the bottom left corner'', ``a pharmacy in the top right corner'').}
\label{fig:city-prompting-based}
\end{figure}

For the second set of experiments, we use our 2D prompting design along with the Flux ControlNet for inpainting (\cref{fig:city-prompting-based}). However, with this setup, the quality of the results is not improved. The layout instructions in the prompt are mostly ignored, again.

Querying Flux to generate large-scale scenes directly has not been successful in our experiments, prompting the need for our grid-based method that allows fine-grained layout and appearance control for each tile.

\subsection{Additional Qualitative Results}%
\begin{figure*}[t]
\centering
\includegraphics[width=0.99\textwidth]{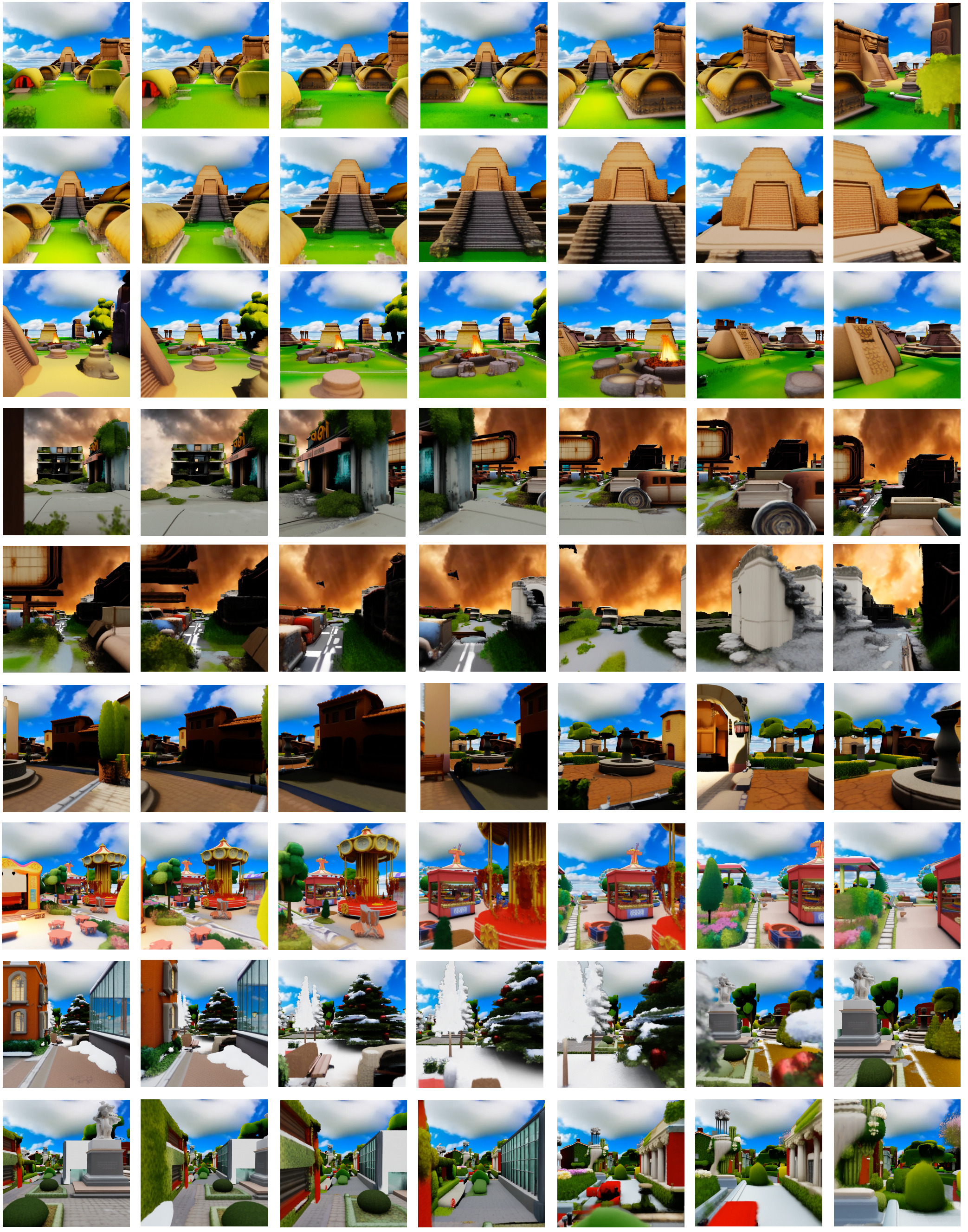}
\caption{\textbf{Exploring a 3D world.} We show trajectories exploring the 3D worlds we generate. A sky box has been added for visual effect.}
\label{fig:explore-app}
\end{figure*}

\begin{figure*}[t]
\centering
\includegraphics[width=0.95\textwidth]{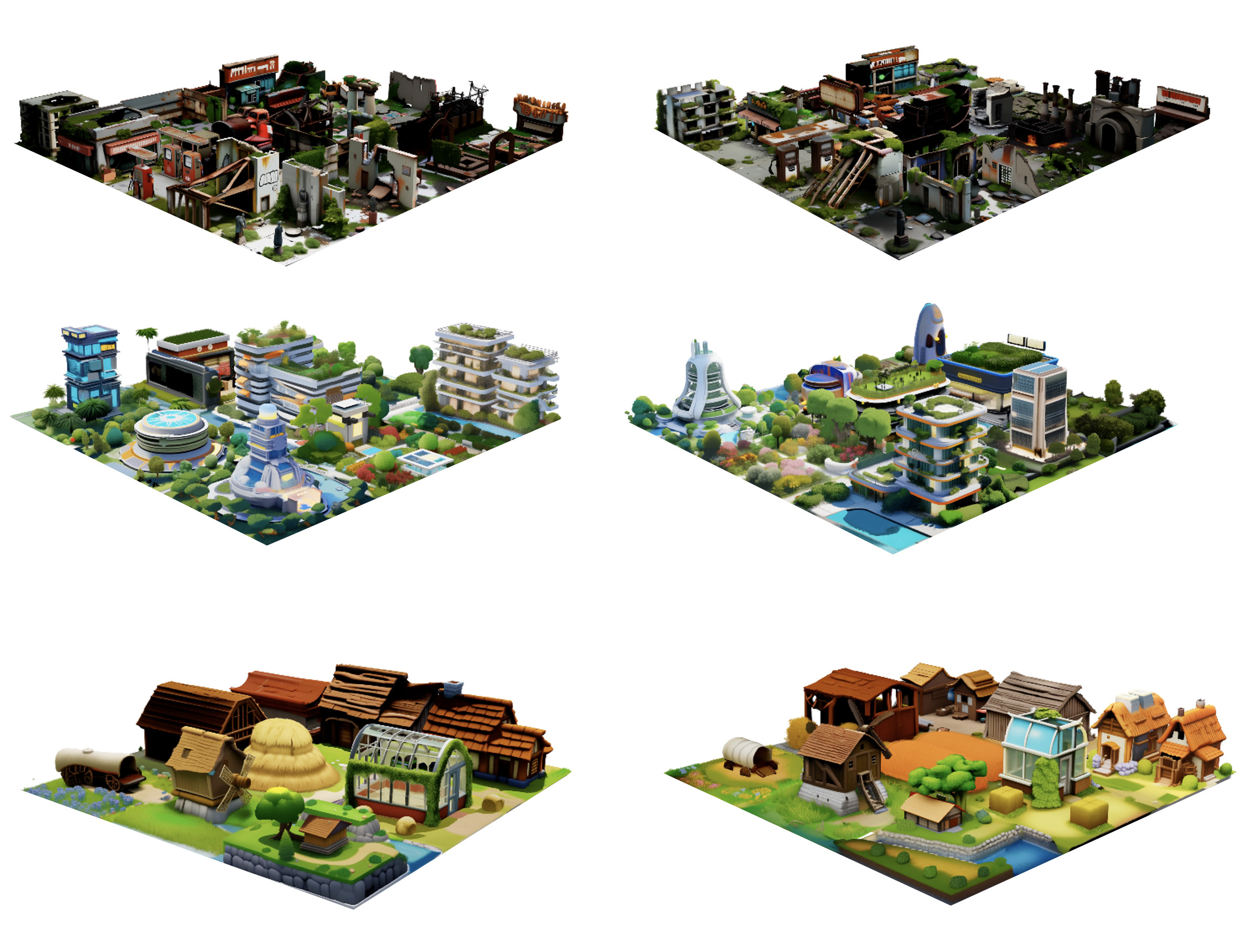}
\caption{\textbf{Generated scenes.} We show scenes generated with the same prompts, but different seeds in 2D inpainting.}
\label{fig:scene1}
\end{figure*}

\begin{figure*}[t]
\centering
\includegraphics[width=0.95\textwidth]{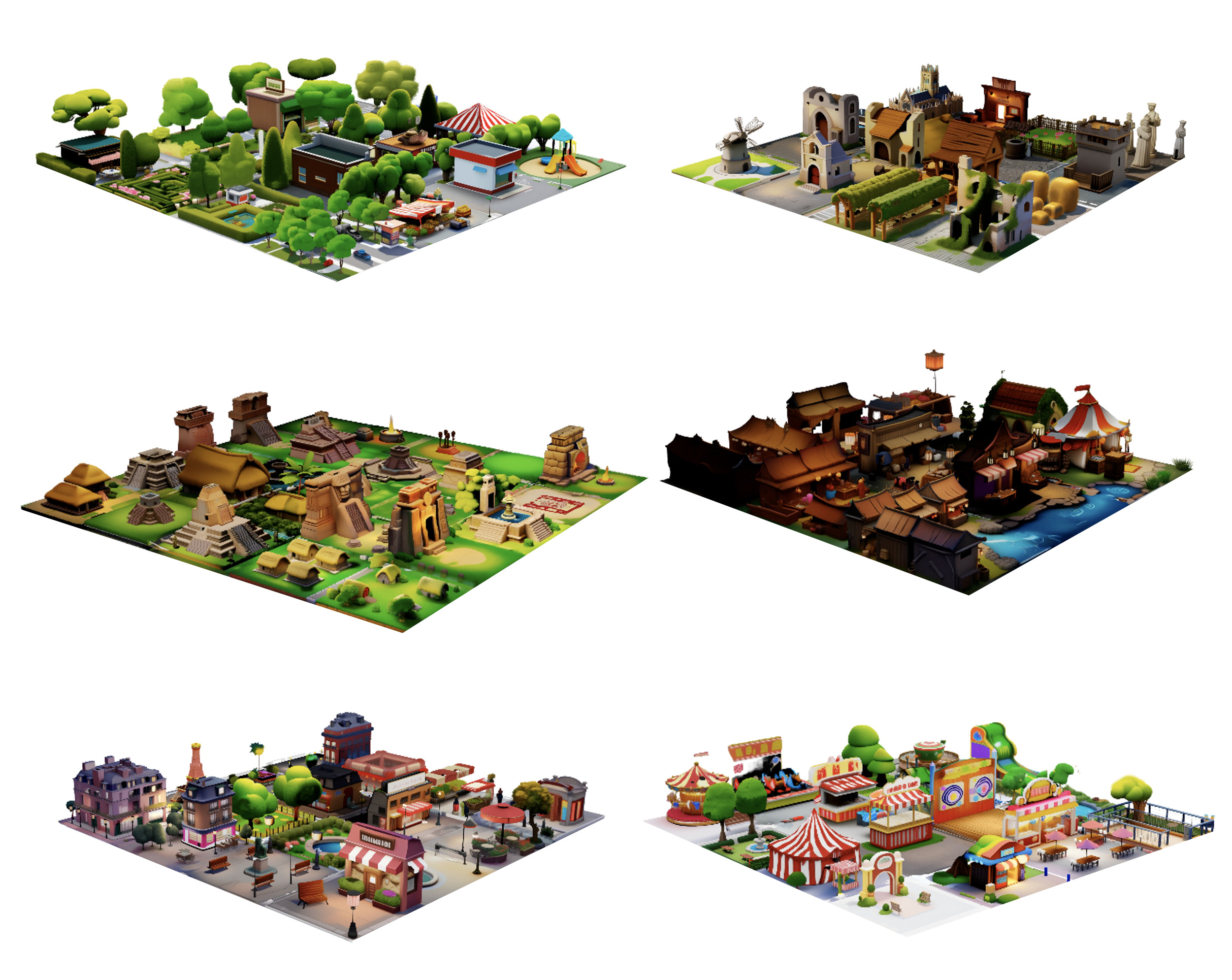}
\caption{\textbf{Generated scenes.}}
\label{fig:scene2}
\end{figure*}

\begin{figure*}[t]
\centering
\includegraphics[width=0.95\textwidth]{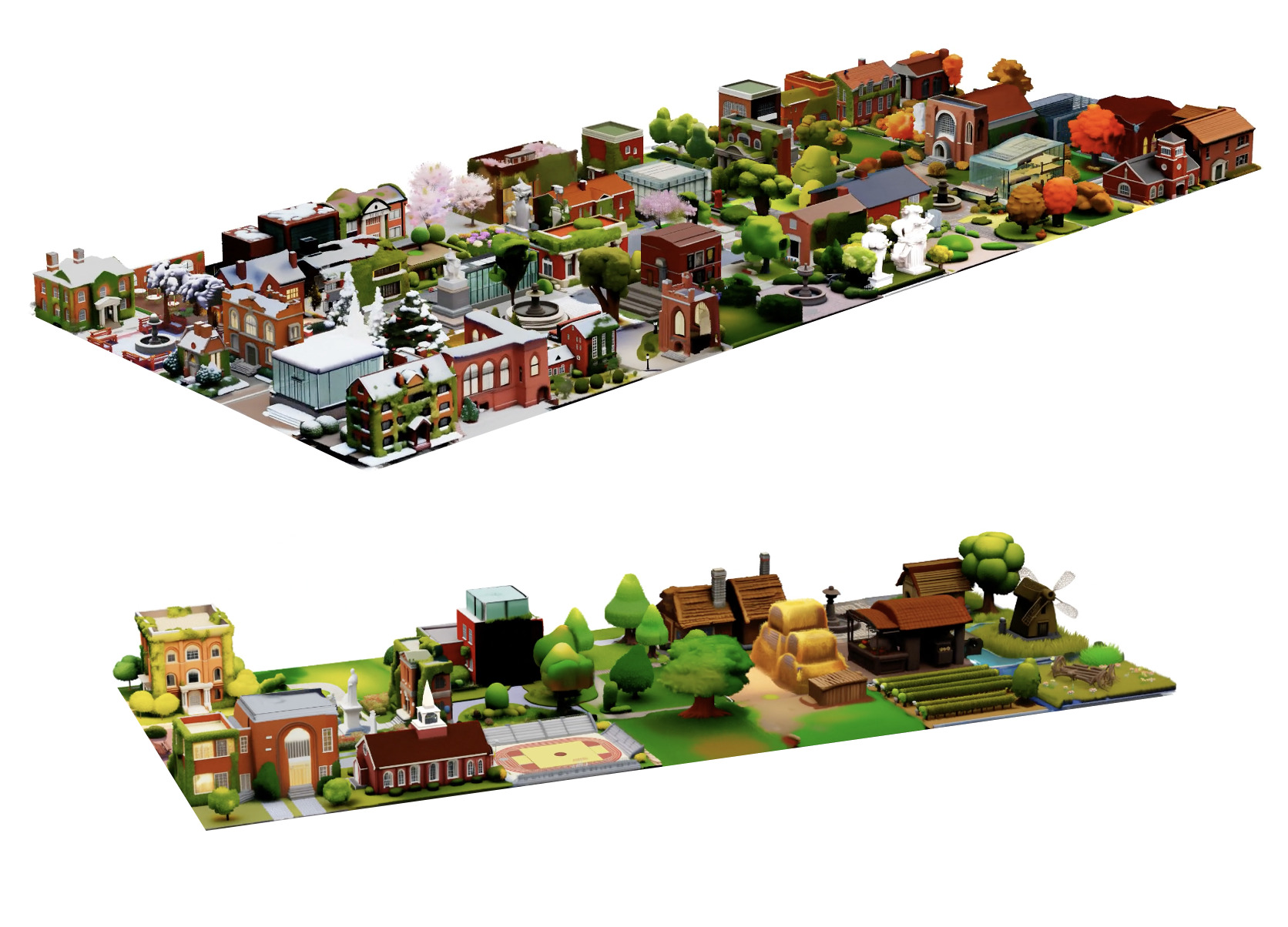}
\caption{\textbf{Generated scenes.} Our method can easily generate large scenes. Further, interesting patterns can be injected thanks to fine-grained control over each tile. \emph{Top:} The scene transitions in season, from winter to spring to summer to autumn. \emph{Bottom:} The scenery transitions from a city-like to a rural environment.}
\label{fig:scene3}
\end{figure*}

In~\cref{fig:explore-app,fig:scene1,fig:scene2,fig:scene3}, we show additional results of our method. As we leverage a pre-trained 2D image generator trained on a very large dataset, we are able to generate highly diverse scenes. Thanks to our fine-grained control at the tile level, we can generate interesting patterns, such as a transition between seasons across a grid (observe the largest grid in~\cref{fig:scene3}).


\subsection{Limitations}%
\label{sec:limitations}
While our method allows creating large and diverse scenes, there are some limitations to be addressed in future work.

\paragraph{Atomic tiles.}
Although we inpaint tiles conditioned on their surroundings, they remain individual units. While structures can be created that span across multiple tiles, this requires harmonious cooperation between Flux and TRELLIS.

\paragraph{Use of heuristics.}
To determine the ground surface height for each tile and removing the base we added during rebasing, we employ heuristics. We have designed these carefully with fallback mechanisms, but they are not infallible.

\paragraph{Inherited limitations.}
As our method builds on top of Flux and TRELLIS, their limitations also apply to ours. During our experiments, we have observed---that despite good inpainting results---TRELLIS at times only vaguely adheres to the conditioning image in terms of appearance, in particular color. Thus, transitions between tiles might not look perfectly smooth (even if they were generated that way in the inpainting result).

\end{document}